\documentclass[acmtog,nonacm]{acmart}

\usepackage{booktabs} %OK % For formal tables

\usepackage{enumitem} %OK
\usepackage{xcolor} %OK
\usepackage{colortbl} %OK

\usepackage[ruled,vlined,noend]{algorithm2e} % For algorithms %OK

\SetCommentSty{mycommfont}

\SetAlFnt{\small}
\SetAlCapFnt{\small}
\SetAlCapNameFnt{\small}
\SetAlCapHSkip{0pt}

\usepackage{mathbbol}
\usepackage{graphicx}
\usepackage{subcaption}
\usepackage{docmute}
\usepackage{listings}
\lstdefinestyle{py}{
  language=Python,
  basicstyle=\ttfamily\footnotesize,
  columns=fullflexible,
  breaklines=true,
  breakatwhitespace=true,
  showstringspaces=false,
  keepspaces=true,
  frame=single,
  tabsize=4
}
\newcommand{\ShowOrHide}[1]{#1} % Shows
\definecolor{darkgreen}{rgb}{0.0, 0.25, 0.0}

\newcommand{\F}[1]{\ShowOrHide{#1}}

\newcommand{\X}{\ensuremath{\mathbf x}}
\newcommand{\Xii}{\ensuremath{\mathbf x_{i}}}
\newcommand{\T}{\ensuremath{\mathbf \tau}}
\newcommand{\Ti}{\ensuremath{\mathbf \tau^{(i)}}}
\newcommand{\Pp}{\ensuremath{\mathbf p}}
\newcommand{\Pj}{\ensuremath{\mathbf p_{j}}}
\newcommand{\Pd}{\ensuremath{f}}
\newcommand{\Pdj}{\ensuremath{f_{j}}}
\newcommand{\A}{\ensuremath{A}}

\begin{document}

% Title portion
\title{Discovering High Level Patterns from Simulation Traces}

\author{Sean Memery}
\email{s.memery@ed.ac.uk}
\orcid{0009-0004-2437-5154}
\affiliation{%
  \institution{University of Edinburgh}
  \city{Edinburgh}
  \country{United Kingdom}
}

\author{Kartic Subr}
\email{k.subr@ed.ac.uk}
\orcid{0000-0002-7302-4383}
\affiliation{%
  \institution{University of Edinburgh}
  \city{Edinburgh}
  \country{United Kingdom}
}

\begin{teaserfigure}
    \centering
    \includegraphics[width=0.93\linewidth]{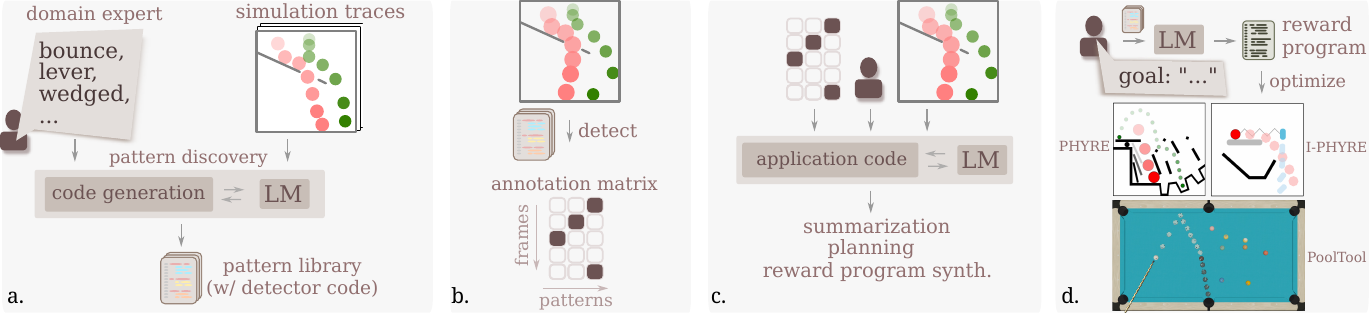}
    \caption{Our method \emph{discovers high-level patterns from low-level simulation traces}, which are useful for Language Models (LMs) to reason about physical systems without fine-tuning. (a) We use evolutionary programming to synthesize programs to detect high-level patterns (e.g., 'ball rolls over obstacle') from raw simulation states. (b) We use a library of such code to translate a simulation trace into a matrix of pattern activations called an annotated trace, which is useful for (c) downstream tasks: \emph{summarization, physics planning, and reward program synthesis}. 
    (d) Using this, we can synthesize reward programs from natural language goals, for physical environments (PHYRE, I-PHYRE, and PoolTool). We optimise these reward programs using traditional methods to solve tasks.
    }
    \Description{}
    \label{fig:teaser}
\end{teaserfigure}

\begin{abstract}
    Large Language Models (LLMs) are unable to reliably reason about specific physical systems. Attempts to imbue LLMs with knowledge of the necessary physics concepts have shown great promise, but explainability and validation remain open challenges.  An emerging alternative is tooling, where LLMs can query physical simulators and use the resulting simulation traces as context for validation. This approach suffers from poor scalability since simulation traces contain large volumes of fine-grained numerical and semantic data. We show that translating simulation traces to a sparse representation of `high-level' structural patterns leads to more effective interpretation by LLMs. We propose an unsupervised learning scheme to perform this translation, or \emph{annotation}, via program synthesis. Our learning results in a library of programs that act as \emph{pattern detectors} which can translate simulation logs to sparse, annotated pattern sequences. The detected patterns may optionally be guided by human experts via string labels (\texttt{rigid collision}, \texttt{stretching spring}, etc.). We show, using a recent physics benchmark, that such annotated representations are more amenable to natural language reasoning about specific physical systems. The synthesised programs serve as transparent, explainable functions that map system states to a sparse and efficient annotation space.  As an example application, we show how goals within physical systems that are specified in natural language may be converted to \emph{reward programs} which are maximised to find solutions.
\end{abstract}




\maketitle

\begin{figure*}[htbp]
    \includegraphics[width=.93\linewidth]{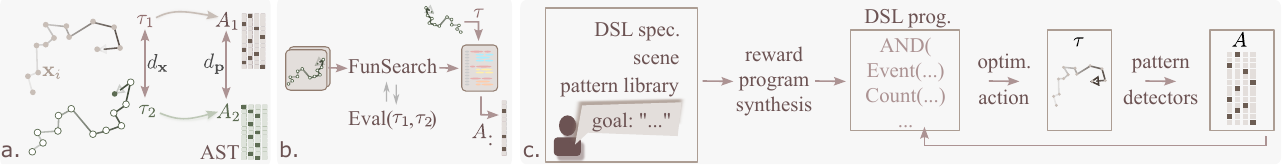}
    \caption{(a) Simulation traces $ \tau_1, \tau_2 $ are mapped to annotated simulation traces (ASTs) $ A_1, A_2$ using detector code. Distance metrics $d_x$ and $d_p$ are defined between traces and ASTs. (b) We use FunSearch \cite{romera-paredesMathematicalDiscoveriesProgram2024}, with a custom evaluation function, to augment the library with new detector code. (c) Given a custom Domain Specific Language, a description of objects in the scene and the current library of pattern-detecting code, we synthesize a reward program in the DSL, which can be optimised to produce actions.
    Simulation traces produced by optimised actions are processed for reward evaluation.
    }
    
    \label{fig:ovw}
\end{figure*}
\section{Introduction}\label{sec:intro}

Imagine a videogame designer assessing feasibility of a level involving physical interaction e.g. a ball needs to bounce against a wall, onto a table and land on a specific target object. This would typically require iteration over level design, simulation and optimisation. Although the optimisation could be trivial if the verification were available as a formal reward program (i.e. code providing verifiable rewards), crafting such programs can be arduous especially where non-trivial interactions are involved. Although descriptions of the goal in natural language would be convenient, converting them into reward programs, while respecting interactions in a specified scene or environment is an open problem. 

An obvious tool to exploit in such problems involving natural language inputs is a Large Language Model (LLM). While current Artificial Intelligence (AI) systems excel at interpreting the goal and in generating plans, they struggle to reliably interpret low-level simulation traces and dynamics~\cite{liuMindsEyeGrounded2022a,mecattafLittleLessConversation2024,xuDeepPHYBenchmarkingAgentic2025, cuetip_paper2025}. Video and multimodal LMs  exhibit limited success on intuitive physics benchmarks~\cite{shivanjassimGRASPNovelBenchmark2023,bordesIntPhys2Benchmarking2025, memery2024simlmlanguagemodelsinfer, xiang2025aligningperceptionreasoningmodeling}. These foundation models remain error-prone and often opaque, offering limited interpretability and explainability beyond the model's own textual justification~\cite{kambhampatiLLMsCantPlan2024, cuetip_paper2025}. Natural language interaction has become a common technique in the graphics community in recent years, with many works utilising language based AI for guiding content generation, LM reasoning, and improved user interaction~\cite{10.1145/3592094, 10.1145/3588432.3591552, 10.1145/3635705, 10.1145/3641519.3657447, 10.1145/3641519.3657422, 10.1145/3641519.3657516, 10.1145/3731209, 10.1145/3721238.3730611}.

Our central observation is that semantic simplification of simulation traces can enable LLMs to forge connections between the specific simulation states and prior knowledge. We propose a method for extracting high-level patterns from raw simulation traces. Given a set of text snippets (e.g.~\texttt{bounce}, \texttt{ball rolling}, etc.) describing potential patterns, we synthesize programs that detect pattern activations from detailed simulation traces. The labels might either be provided by a domain expert (game designer in our motivating example) or an LLM. We learn a program corresponding to each label, to detect the occurrence of that pattern in a simulation trace. 
%
% We demonstrate that LLMs are able to effectively interpret these high-level annotated traces to generate summaries, perform physics reasoning tasks, and to automatically synthesize formal and interpretable reward functions from natural language goals.

\begin{figure*}[t]
    \includegraphics[width=\linewidth]{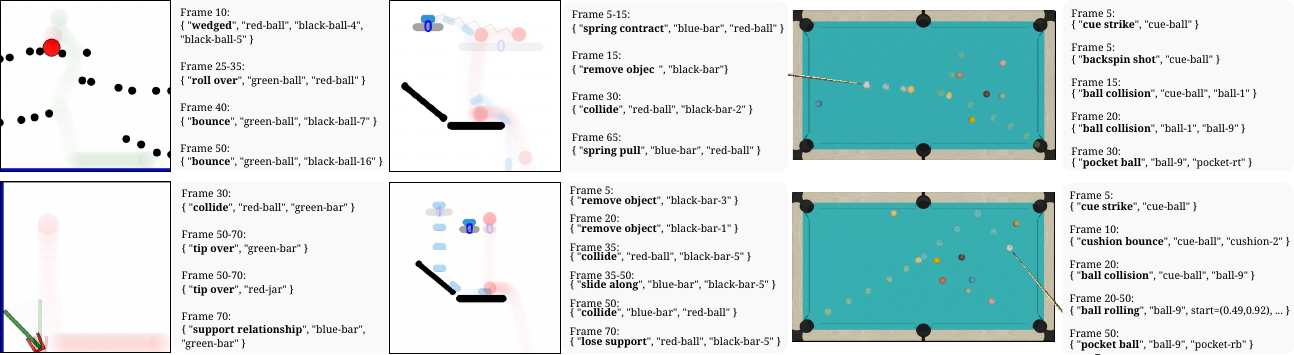}
    \caption{Examples of annotations produced via Human-Label libraries on two scenes each (rows) from PHYRE, I-PHYRE, and PoolTool environments (columns). 
    % Rendered frames and time-stamped pattern activations are shown for each scene.
    }
    
    \label{fig:pattern_examples}
\end{figure*}

The \emph{library of pattern-detecting programs} generalizes across scenes but is learned individually for each environment\F{, i.e. we do not achieve cross-environment transfer but focus on environment specific learning}. We evaluate the effectiveness of libraries across thousands of scenes within three different environments available within the DeepPHY~\cite{xuDeepPHYBenchmarkingAgentic2025} benchmark. We learn programs by extending FunSearch~\cite{romera-paredesMathematicalDiscoveriesProgram2024} and without the need for a supervisory dataset. 
We use the library to translate low-level simulation trace data into \emph{Annotated Simulation Traces} (ASTs) containing high-level pattern sequences identifiable via natural language labels. Figure~\ref{fig:pattern_examples} shows examples from each environment. We show that, with ASTs as input, LLMs are more effective at summarization, solving reasoning tasks, and synthesizing reward programs from natural language goals. 
In summary, our contributions are:
\begin{itemize}[noitemsep, leftmargin=1.5em, topsep=.1em]
    \item we introduce the concept of extracting high-level patterns from simulation traces, for applications involving natural language reasoning in environments with physical interaction;  
    \item we invent a method that relies on minimal (and optional) user-guidance string labels to discover patterns from simulation data;
    \item we use evolutionary program synthesis to learn interpretable pattern detector programs for annotating simulation traces;
    \item and we show the effectiveness of annotated traces using physics problem solving, summarization, and reward program synthesis. 
\end{itemize}

\section{Related work}

\subsection{Language models and physics environments}
There has been much interest in whether human intuitive physics is driven by approximate ``intuitive physics engines'' that support fast counterfactual prediction under uncertainty~\cite{battagliaSimulationEnginePhysical2013}. This is relevant to interactive AI agents operating in physics environments. 
Verbal interaction and reasoning about physics naturally suggests exploitation of language models (LMs) and vision-language models (VLMs). Recent work evaluates whether modern LMs and VLMs acquire comparable physical priors from large-scale pretraining, using both text-centric and video-centric benchmarks. 

Curated physics problem sets involving text and/or visual data as context is prominent ~\cite{qiuPHYBenchHolisticEvaluation2025, Xu2025PhySensePP, Chow2025PhysBenchBA, Xiang2025SeePhysDS, Zhang2025PhysReasonAC, Dai2025PhysicsArenaTF}. However simulation-based benchmarks probe intuitive physical principles under temporal dynamics, some examples include GRASP~\cite{shivanjassimGRASPNovelBenchmark2023} and IntPhys2~\cite{bordesIntPhys2Benchmarking2025}, which evaluate vision-language models intuitive physics ability. Complementary evaluation suites target agentic performance in dynamic environments and games (e.g., BALROG~\cite{paglieriBALROGBenchmarkingAgentic2024}) and embodied interaction benchmarks (e.g., LLM-AAI~\cite{mecattafLittleLessConversation2024}) as well as explicitly physics-focused agentic VLM evaluation (e.g., DeepPHY~\cite{xuDeepPHYBenchmarkingAgentic2025}). Across these settings, results generally indicate that strong language-based reasoning does not translate into robust physical prediction and control, and rarely approaches human level.

A common response is to ground reasoning in external tools or simulators. Mind’s Eye~\cite{liuMindsEyeGrounded2022a} conditions an LM on outcomes generated by a physics simulation to improve physics question answering. 
It has also been shown that simulation can be used in closed-loop to ground LM reasoning~\cite{cuetip_paper2025, cherianLLMPhyComplexPhysical2024a}. Several works argue that LMs are most reliable when paired with external model-based verifiers rather than used as standalone planners~\cite{memery2024simlmlanguagemodelsinfer, kambhampatiLLMsCantPlan2024, cuetip_paper2025}. In parallel, there is evidence that improving visual representations and training signals may be necessary for physics understanding. This includes the V-JEPA series of models ~\cite{assran2025vjepa2selfsupervisedvideo, garridoIntuitivePhysicsUnderstanding2025a}, where predicting physics outcomes within a learned representation space has shown promise. Additionally, reinforcement learning of vision-LMs in synthetic worlds has been shown to improve 3D embodied behavior~\cite{bredisEnhancingVisionLanguageModel2025}. 

We build on two recent insights: First, that LMs are more effective when reasoning about high-level events rather than low-level simulation state traces~\cite{cuetip_paper2025,memery2024simlmlanguagemodelsinfer}; and second, that LMs are effective at generating executable code that models environment dynamics and structure~\cite{tang2024worldcodermodelbasedllmagent,daineseGeneratingCodeWorld2024, romera-paredesMathematicalDiscoveriesProgram2024}. The latter learns executable transition models from interaction data.
Inspired by these works, we learn code to detect high-level patterns from simulation traces, and use these patterns to support LM reasoning about physical systems.

% CueTip~\cite{cuetip_paper2025} instruments a physics simulator to emit natural-language event traces alongside state traces and uses these events to produce explanations grounded in expert-provided guidelines. We build on the core idea of event-based LM interfaces, but replace expert rules with a learned library of pattern detectors, and use the resulting events not only for explanation but also for benchmarking (Q\&A), summarization, and reward program synthesis. We are also complementary to code-based world modeling approaches such as WorldCoder~\cite{tang2024worldcodermodelbasedllmagent} and Code World Models work~\cite{daineseGeneratingCodeWorld2024}, which aim to learn executable transition models through interaction; in contrast, we learn executable detectors that annotate simulation traces.

\subsection{Reward program synthesis}
Classical approaches to inverse reinforcement learning aim to infer rewards explaining expert behavior \cite{ng2000irl}. Some methods recover structure and interpretability, such as specifications with temporal structure \cite{NEURIPS2018_74934548} or symbolic reward representations such as reward machines~\cite{toro2018reward,NEURIPS2019_532435c4}. We are inspired by reward learning via program synthesis, inducing interpretable reward programs by example~\cite{Zhou_Li_2022}, demonstrations or preferences. Eureka~\shortcite{Ma2023EurekaHR} uses LLMs to synthesize RL reward functions from task descriptions, iteratively refining them based on performance feedback. While effective at learning useful reward functions, this lacks interpretability and compositional structure. 
 
Reward programs can be optimised to produce controllable behavior~\cite{Davidson_2025,yu2023languagerewardsroboticskill} and are closely related to model-based reasoning and planning via program learning with strong compositional structure~\cite{curtis2025llmguidedprobabilisticprograminduction,tang2024worldcodermodelbasedllmagent,ahmed2025synthesizingworldmodelsbilevel}.
% Probabilistic-programming perspectives treat cognition and reasoning as the synthesis of task-specific generative models. For example, 
Open-world cognition may be modeled as iterative construction and refinement of probabilistic models~\cite{wong2025modelingopenworldcognitionondemand}.
The above methods use program synthesis to explicitly make goals, models, and evaluators, allowing adaptability, interpretability and external verification. 

% LMs have been shown to be effective at generating candidate proposals for reward programs.  

\subsection{Program synthesis via FunSearch}
FunSearch~\cite{romera-paredesMathematicalDiscoveriesProgram2024} is a method for genetic programming (GP) and hence a form of evolutionary algorithm (EA). It retains the general loop structure of EAs, which is to maintain a population of candidate solutions and iteratively apply variation and selection to evolve better solutions. It searches the functional space of executable programs by replacing traditional rule-based or stochastic changes with a large language model (LLM) to handle mutation and discovery. Hallucination is controlled by an execution-based evaluation function that scores candidates within domain-specific contexts. It employs an `island model' where multiple sub-populations evolve in parallel with occasional migration of high-performing candidates between islands to maintain diversity.
Related methods incorporate evolutionary search or explicit reflective feedback to improve sample efficiency and exploration, including Evolution of Heuristics (EoH)~\cite{liuEvolutionHeuristicsEfficient2024} and ReEvo~\cite{yeReEvoLargeLanguage2024a}. Tree-search variants such as GIF-MCTS~\cite{daineseGeneratingCodeWorld2024} further combine LM proposals with structured exploration to generate reliable code for environment modeling and planning. 

We adapt the general scheme (algorithm in appendix F) to synthesize pattern detectors. Rather than optimising towards a single labeled output, we score candidate programs by whether their emitted pattern streams covary with meaningful differences in trace geometry, while discouraging redundancy with respect to the current library (Sec.~\ref{sec:patterndiscovery}).
We achieve this by providing (i) a user-defined evaluation function $ \texttt{evaluate}(\cdot) $ to score candidate outputs (and reject invalid ones). See appendix E for LM prompts used with FunSearch.

\section{Method}\label{sec:method}
\subsection{Definitions: Patterns, annotations,  detectors, distances}
Let $\Xii \in \mathcal{X}$ be the state in the state space of the physics environment at the $i^{th}$ time-step of the simulation and $ \T = \{\X_1, \X_2, \ldots, \X_N\} \in \Upsilon$ be a simulation trace of length $|\T| = N$. Let $\T_i$ denote the $i^{th}$ trace while $\Ti$ represents the $i^{th}$ step of \T\ (\Xii\ in this case).
We define an alternative, abstract space for simulation traces to enable high-level reasoning involving common patterns. 

A \emph{pattern} $\Pp \in \mathcal{P}$ captures a specific evolution of states within \T. For example, a subsequence of states representing elastic collision. These subsequences are not mutually exclusive across states and so \T\ cannot be strictly defined as a sequence of patterns. We use a $|\T| \times |\mathcal{P}|$ sparse binary \emph{annotation matrix}, $\mathcal{P}$ is the \emph{pattern library} and $\A_{ij} = 1$ iff the $j^{th}$ pattern $\Pj$ is active at time-step $i$.

For each pattern \Pj\ in the library, we define a \emph{pattern detector} as a program that acts as a function $\Pdj: \Upsilon  \times \Theta \rightarrow [0,1]^N$. In the above example, $\Pdj(\T, \theta)$ outputs the $j^{th}$ column of the annotation matrix, $\A_{:j}$ where $\theta_j \in \Theta$ represents pattern-specific parameters. For example, $\theta$ could contain identification numbers of objects involved in the pattern. Also associated with each pattern is a label $L_j$ which is a short natural language description of the pattern (e.g., ``elastic collision between objects X and Y'').

We define two distance metrics to compare traces in the state and annotation space respectively. We define \emph{trace distance} $d_\X$, as a normalized translational distance of matched objects. That is, if $O$ is the intersection of the sets of objects present in traces $\T_1$ and $\T_2$ respectively, then $d_{\X} (\T_1, \T_2)$ measures the average Euclidean distance between each object in $O$ per frame, across $\T_1$ and $\T_2$, normalized by the length of the trace.

We define \emph{pattern annotation distance} $d_\Pp$, as the cross entropy between normalized histograms of pattern occurrences over time, averaged across all patterns in the library. That is, given annotation matrices $\A_1$ and $\A_2$ and a pattern library with $J$ patterns, we discretize each into $b$ bins and compute normalized histogram counts $\mathbf h_j(\A_1, b)$ and $\mathbf h_j(\A_2, b)$ for $j=1,2,\cdots,J$. Then, $d_{\Pp} (\A_1, \A_2)$ is the cross entropy between $\mathbf h_j(\A_1, b)$ and $\mathbf h_j(\A_2, b)$, averaged across $J$ patterns. This distance captures how similar the distributions of activations of patterns are across annotations of the two traces.

\subsection{Natural language guided pattern discovery}\label{sec:patterndiscovery}
Given a set of traces $\{\T_k\}_{k=1}^K$, we learn a library of pattern detectors $\mathcal{P} = \{\Pdj\}_{j=1}^J$ that discover patterns from traces of simulation states ${\X_1, \ldots, \X_N}$ across scenes within an environment. We score candidate pattern detectors based on fidelity and novelty, and add penalty terms to discourage long programs, degenerate patterns and slow execution times (Appendix F).

% (1) the correlation between trace distance and the obtained pattern annotation distance, and (2) information gain provided relative to existing patterns in the library. 
% We introduce penalty terms to discourage long programs, degenerate patterns and slow execution (Appendix F).  

Our primary consideration for fidelity is that patterns should reflect similarities between simulation traces. That is for two traces $\T_1$ and $\T_2$, with corresponding annotation matrices $\A_1$ and $\A_2$, if the traces are similar then their pattern annotations should also be similar, and vice versa. In other words, a high correlation between the distances in those spaces, $d_\X(\T_1, \T_2)$ and $d_\Pp(\A_1, \A_2)$, is desirable.
In addition, we seek to discover patterns that are informative with respect to the existing library. For a candidate pattern detector $\Pd_{new}$ producing annotation matrix $\A_{new}$, we want 
$d_\Pp(\Pdj, \Pd_{new})$ to be high, encouraging novelty. These desirables are achieved by incorporating them in the fitness function ($\rho$ and $\eta$, respectively, in Alg.~\ref{alg:compute_fitness})

We start from a pool of candidate pattern labels (natural language) provided by a user and use an evolutionary programming approach to search for corresponding pattern detectors that maximize the above fitness criteria. The pattern discovery algorithm (Algorithm~\ref{alg:pattern_discovery}) takes as input a set of traces $\mathcal T$, a set of candidate pattern labels $\mathcal L$ and a skeleton (seed) program $g_0$ which contains empty logic with the structure required of a pattern detector. After  initializing $\mathcal{P}$ and setting up some parameters, it invokes FunSearch to synthesize a candidate pattern detector  $\Pd_{m}^*$ for each $L_m \in \mathcal L$ (along with its fitness score $\nu_m^*$). If $\nu_m$ exceeds a predefined threshold $\delta$, the candidate pattern detector is added to the library $\mathcal{P}$. 

\begin{algorithm}[h]
\SetAlgoLined
\KwIn{ Set of traces $\mathcal T = \{\T_k\}_{k=1}^K$, \\ 
        \hspace{2.7em}  candidate pattern labels $\mathcal L = \{L_m\}_{m=1}^M$. \\
        \hspace{2.7em}  skeleton algorithm $g_0$}

\KwOut{Pattern library $\mathcal{P} = \{\Pdj\}_{j=1}^J$}
 Initialize empty pattern library $\mathcal{P} \leftarrow \{\}$\;
 Initialize LLM for FunSearch $\texttt{LLM}(\cdot)$\;
 Initialize FunSearch parameters $I, s, T_r$ \;
 \For{each label $L_m$ in $\mathcal L$}{
   $(\Pd_{m}^*, \nu_m^*) \gets 
   \texttt{FunSearch}( \mathrm{Evaluate( \mathcal{T}, \; \cdot, \;  \mathcal{P})}, \; g_0,  \;   \texttt{LLM},  \;  I, s, T_r)$\;
   \If{$\nu_m^* > \delta$}{
      $\mathcal{P} \leftarrow \mathcal{P} \cup \{\Pd_{m}^*\}$\;
   }
 }
 \caption{DiscoverPatternDetectors}
 \label{alg:pattern_discovery}
\end{algorithm}

FunSearch (Algorithm~\ref{alg:pattern_discovery}) uses the Evaluate function (Algorithm~\ref{alg:compute_fitness}) as the fitness function for candidate pattern detectors. Given a set of traces $\mathcal T$, a candidate pattern detector $\Pd_\mathrm{new}$ and the current library $\mathcal{P}$, it computes the trace distances $d_\X$ and pattern annotation distances $d_\Pp$ for all pairs of traces in $\mathcal T$ using the current library and the candidate pattern detector. It also computes distances between annotations by the new pattern and annotations by the existing library, where a higher mean distance indicates greater novelty. Finally, it computes the correlation $\rho$ between $D_\X$ and $D_\Pp$, novelty score $\eta$, length penalty $\lambda$ and time penalty $\psi$, and combines them to produce the final fitness score $\nu$. Parameters $\theta$ for each pattern detector are inferred during synthesis and output as metadata by the synthesized program.

\begin{algorithm}[h]
\SetAlgoLined
\KwIn{Set of traces $\mathcal T = \{\T_k\}_{k=1}^K$, \ 
        \hspace{2.7em}  Candidate pattern detector $\Pd_\mathrm{new}$, \
        \hspace{2.7em} Current library $\mathcal{P}$}
\KwOut{Fitness score $\nu$}
 Initialize empty lists $D_\X \leftarrow []$, $D_\Pp \leftarrow []$, $D_\mathrm{novel} \leftarrow []$\;
 \For{each pair of traces $(\T_l, \T_m)$ in $\mathcal T$}{
    Compute annotation matrices $A_l$, $A_m$ using current library $\mathcal{P}$\;
    Compute annotation vectors $\mathbf a_l$, $\mathbf a_m$ using $\Pd_\mathrm{new}$\;
    % Compute trace distance $d_\X(\T_l, \T_m)$\;
    % Compute pattern annotation distance $d_\Pp(\mathbf a_l, \mathbf a_m)$\; 
    Append $d_\X(\T_l, \T_m)$ to $D_\X$\;
    Append $d_\Pp(\mathbf a_l, \mathbf a_m)$ to $D_\Pp$\;
    Append $d_\Pp(\mathbf a_l, A_l)$ and $d_\Pp(\mathbf a_m, A_m)$ to $D_\mathrm{novel}$\;
 }
 $\rho \gets \mathrm{corr}(D_\X, D_\Pp)$\;
 $\eta \gets \mathrm{mean}(D_\mathrm{novel})$\;
 $\lambda \gets \mathrm{ComputeLengthPenalty}(\Pd_\mathrm{new})$\;
 $\psi \gets \mathrm{ComputeTimePenalty}(\Pd_\mathrm{new})$\;
 Compute fitness score: $\nu \gets \rho + \eta - \lambda - \psi$\;
 \caption{Evaluate}
 \label{alg:compute_fitness}
\end{algorithm}

\subsection{Optimised program ensembles per pattern}
We robustify the library via multiple independent executions of the discovery procedure  with different random seeds, yielding a collection of learned libraries $ \mathcal{P}_1, \mathcal{P}_2, \ldots, \mathcal{P}_M $. 
%i.e. the $j^{th}$ pattern (with label $ L_j$) is associated with a set of $M$ candidate detectors $\{\Pd_{j,1}, \Pd_{j,2}, \ldots, \Pd_{j,M}\}$, one in each library. We cluster these candidates (described below) and represent each pattern $L_j$ in the final library as a weighted sum of detectors, one per cluster. While this adds extra computation time during learning, the extra cost at inference time is negligible, since the number of clusters is small and the detectors are simple to execute. Section~\ref{sec:ensemble_ablation} shows that our method may be used even without this step, at the cost of some performance reduction.
We cluster the entire pool of code candidates from these runs and represent each pattern $L_j$ in the final library as a weighted sum of detectors, one per cluster. We do this by applying K-means clustering to the code candidates based on their pattern annotation distances $d_\Pp$. While this adds extra computation time during learning, the extra cost at inference time is negligible, since the number of clusters is small and the detectors are simple to execute. Section~\ref{sec:ensemble_ablation} shows that our method may be used even without this step, at the cost of some performance reduction.

\paragraph{Pattern activation clustering} Given a trace $\T_k$, we execute the ensemble of pattern detectors for the $j^{th}$ label, $\Pd_{j,\ell}$  and obtain an $N\times K$ annotated trace matrix $\hat{A}$ where $\hat{A}_{i,\ell} \in [0,1]$ represents the activation level of the $\ell^{th}$ detector for the $i^{th}$ time-step. We cluster the columns of $\hat{A}$ using spatiotemporal tolerances $(\epsilon_s, \epsilon_t)$ (where the spatial component is based on the distance between the objects and the temporal component is based on the timestep of each activation) and on whether they share the same parameters $\theta$. A weighted average is used to calculate the activation of each cluster.  

\paragraph{Reliability weighting and tolerance optimisation} The reliability weights $w_{j,\ell}$ of the $j^{th}$ pattern detector are initialized to unity and refined via three steps, along with the spatiotemporal tolerances. First, we compute the activation of cluster $C$ as $a_C = \sum_{\ell \in C} w_{j,\ell} \; t_{j,\ell}$ where $t_{j,\ell}$ is the  training reward obtained for the $\ell^{th}$ detector. If $a_C>\gamma$, we accept the cluster as having been activated. Second, we perform a hyperparameter sweep to maximize the correlation-based reward on held-out data. Finally, we refine the weights via Bayesian optimisation. We sample 20 traces from the environment dataset and compute their annotations using the ensemble library. Each annotation is then summarized into natural language. For each visualized trace, we present a language model with (i) images of that trace and (ii) 8 sampled summaries, of which exactly one is the matching summary for the trace. The model is asked to identify the best matching summary. The resulting identification accuracy is used as the reward signal for Bayesian optimisation over the detector weights and tolerances. In this way, the ensemble is calibrated directly against downstream interpretability and discriminative utility, rather than detector reward alone. \F{The spatiotemporal tolerances for each environment are: PHYRE $(\epsilon_s, \epsilon_t) = (0.05, 4)$, I-PHYRE $(\epsilon_s, \epsilon_t) = (0.00625, 2)$ and PoolTool $(\epsilon_s, \epsilon_t) = (0.005, 1)$.}

In summary, at the end of this step the robustified pattern library $\mathcal{P}^*$ contains a set of programs per pattern, whose weighted sum determines whether that pattern is activated. 

\subsection{Reward program synthesis}\label{sec:rewardsynthesis}

%Here we describe how agents may be optimised to achieve high-level goals specified in natural language. 
Given a natural language goal (e.g., ``make the red ball collide with the green object'' for PHYRE or ``knock the green ball into the red ball'' for PoolTool), we synthesize a compositional expression in a custom  domain-specific language (DSL) and call it a \emph{reward program}.
The reward program operates on an \emph{annotated simulation trace} (AST) to produce (1) a boolean success/failure signal and (2) a dense reward signal in $[0,1]$ indicating partial credit towards goal completion. The program can then be used as a reward function for trace optimisation. 

\begin{figure*}
    \includegraphics[width=.85\linewidth]{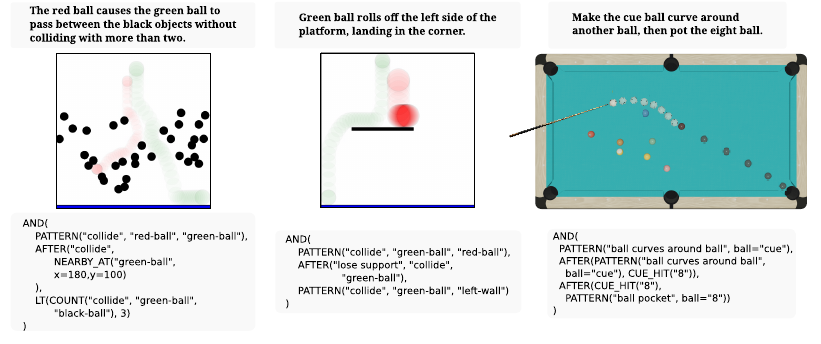}
    \caption{Three examples (columns) of optimised actions (middle row) using DSL programs (bottom row) synthesized by our method.}
    
    \label{fig:dsl_examples}
\end{figure*}

\begin{figure*}
    \includegraphics[width=.93\linewidth]{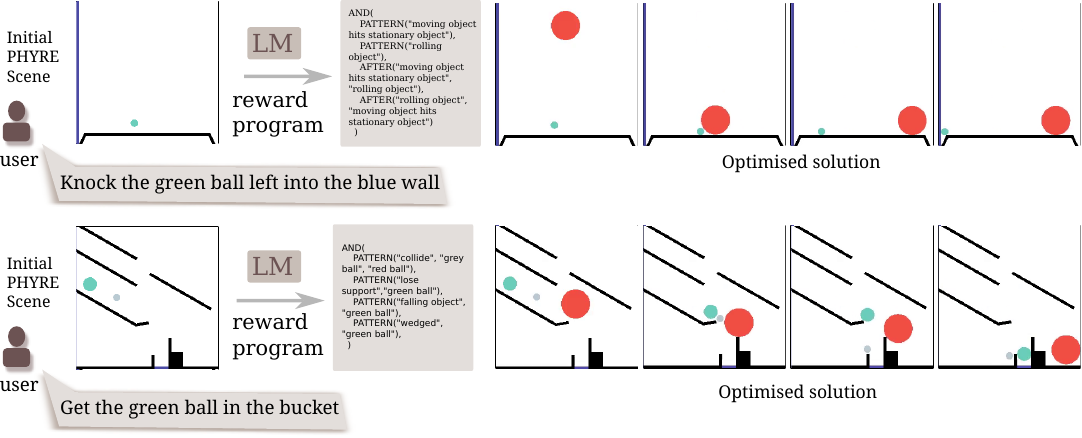}
    \caption{Two examples of PHYRE reward programs being optimised for NL goals.}
    
    \label{fig:phyre_reward_program}
\end{figure*}

\begin{figure*}
    \includegraphics[width=.93\linewidth]{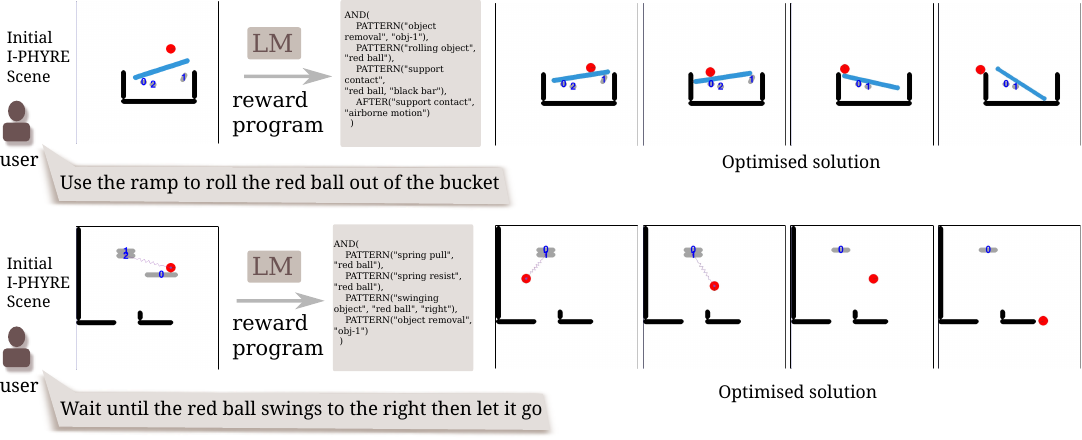}
    \caption{Two examples of I-PHYRE reward programs being optimised for NL goals.}
    
    \label{fig:iphyre_reward_program}
\end{figure*}

\begin{figure*}
    \includegraphics[width=.93\linewidth]{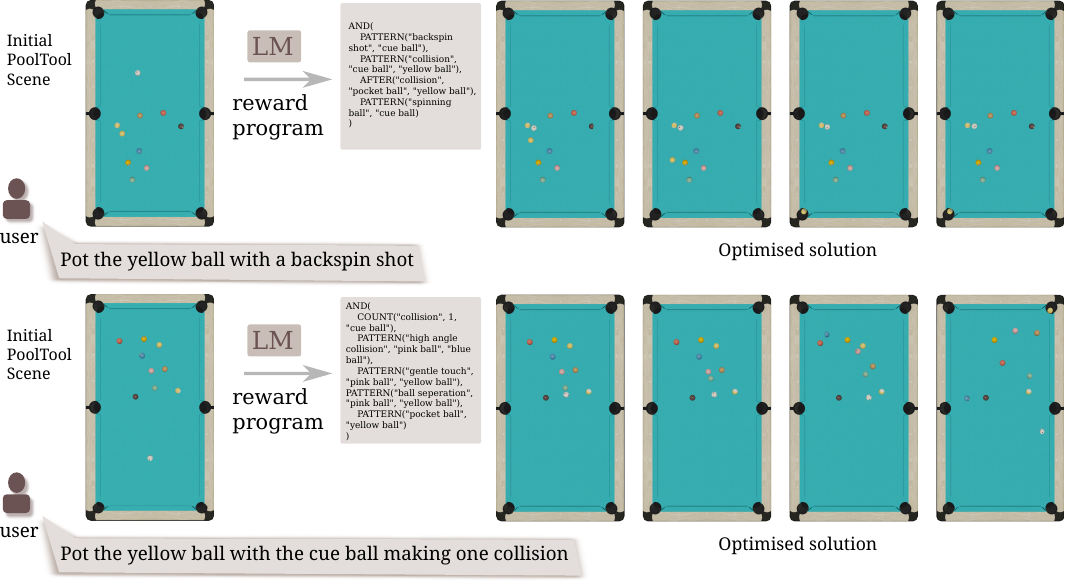}
    \caption{Two examples of PoolTool reward programs being optimised for NL goals.}
    
    \label{fig:pooltool_reward_program}
\end{figure*}

We prompt the language models with a structured prompt denoted $P_r(G, \mathcal{P}, \Theta, D, \{(\hat{G}_k, \hat{r}_k)\})$ where $G$ is the natural language goal, $\mathcal{P}$ is the pattern library with associated parameter sets $\Theta$, $D$ is a description of the DSL syntax and semantics, and $\{(\hat{G}_k, \hat{r}_k)\}$ is a set of few-shot examples of natural language goals and their corresponding reward programs. Since the synthesized programs may contain syntax errors, invalid identifier usage or mismatched parameter keys, if parsing or execution fails we iteratively prompt for automatic repair by supplying the candidate DSL program and the interpreter error message. We abort as a failure after a fixed retry limit is reached. See Appendix B for the full list of learned patterns, and Appendix E for LM prompts. Figure~\ref{fig:dsl_examples} shows optimised actions for synthesized programs, with additional examples from PHYRE, I-PHYRE, and PoolTool in Figures~\ref{fig:phyre_reward_program}--\ref{fig:pooltool_reward_program}.
The reward program $r(\cdot)$ is structured as a single DSL expression composed of multiple boolean predicates. When executed on an AST containing tuples of the form $\left( L_j, i, \theta_j \right)$ (where $L_j$ is a pattern label active at time-step $i$ with parameters $\theta_j$), $r$ serves as a test for whether the natural language goal $G$ was achieved. We use three classes of predicates and one quantitative primitive. \emph{Pattern predicates} check for the occurrence of specific patterns within a trace. \emph{Logical predicates} facilitate classical boolean operators such as AND, OR and NOT.  \emph{Temporal predicates} test activation timings and relative ordering of patterns in the trace. \emph{Spatial and frequency quantifiers} measure spatial proximity and frequencies.

\paragraph{Partial-credit scoring}
In addition to boolean satisfaction, we compute a dense reward in $ [0,1]$. We interpret the synthesized program to be composed of a top-level AND operator with  multiple operands and return a reward that is the average of the number of subclauses that evaluate to true. For quantitative primitives, we assign graded scores based on the distance-to-satisfaction. For \texttt{NEARBY\_AT}, we convert the object-to-target distance into a score using an inverse-log transform and clamp to $ [0,1] $, so improvements near the target are weighted strongly. For \texttt{COUNT} and comparisons, we compute a deviation from the target count and map it to $ [0,1] $ again with an inverse-log shaping and clamping. 
For example, the goal ``Pot the 9-ball in the lower left pocket without touching a cushion'' may be synthesized into the following reward program:
{\footnotesize
\begin{verbatim}
AND(
  # Curve the cue ball around the obstacles
  PATTERN(
    "ball curves around ball",
    {"object_a": "cue-ball", "object_b": "black-ball"}
  ),
  # Collide with the 9-ball after the curve
  AFTER(
    "ball collision", "ball curves around ball",
    first_params={"object_a": "cue-ball", "object_b": "9-ball"},
    second_params={"object_a": "cue-ball", "object_b": "black-ball"}
  ),
  # Ensure that the 9 ball does not touch any cushions
  NOT(PATTERN("cushion rebound")),
  # 9-ball get pocketed in the correct pocket
  PATTERN(
    "ball pocketed",
    {"object_a": "9-ball", "object_b": "pocket-lower-left"}
  ),
  # End with the cue ball in a beneficial position
  NEARBY_AT(
    "cue-ball", x=0.25, y=1.15, t=1.0
  ),
)
\end{verbatim}
}
Here, we check for the existence of a ``ball curves around ball'' pattern involving the cue ball. The ``AFTER'' call enforces that the ``ball collision'' pattern involving the cue ball and the 9-ball must occur after the first pattern activation. Finally, the \texttt{NEARBY\_AT} predicate checks if the cue ball is near the specified coordinates at the end of the trace, which can target the desired position.

\section{Evaluation of learned patterns}\label{sec:experiments}

This section presents validation of our method for learning pattern libraries. We  describe the evaluation benchmark (Section~\ref{sec:benchmark}), assess the differences between human- and LLM-labeled libraries (Section~\ref{sec:human_vs_llm}) and show the effect of library size on downstream performance (Section~\ref{sec:library_size}). We also present some results from a user study (Section~\ref{sec:user_study}).
\F{We use Qwen3.6 35B A3B to learn all libraries, and evaluate them with a range of open-source Qwen vision-language models of different sizes \cite{bai2025qwen3vltechnicalreport,qwen36_35b_a3b}, using \texttt{LLaMA.cpp} \cite{llamacpp} as the inference backend.} We provide more qualitative examples of learned detectors, ensemble annotations, generated summaries, and reward programs in appendix G and H. These examples are intended to illustrate both the strengths and the failure modes of the learned libraries.

\subsection{Evaluation benchmarks and tasks}\label{sec:benchmark}

We evaluated our method on hundreds of scenes set up within three different physics simulation environments in DeepPHY~\cite{xuDeepPHYBenchmarkingAgentic2025}: PHYRE~\cite{bakhtin2019phyrenewbenchmarkphysical}, I-PHYRE~\cite{li2024iphyreinteractivephysicalreasoning}, and PoolTool~\cite{Kiefl2024}, with the same environment-level benchmarks and tasks as DeepPHY. We learned an ensemble library of $12$ patterns from $M=10$ independent runs for each environment. 
% We therefore refer the reader to DeepPHY for full details of the environments and benchmark protocols. Our contribution here is not a new benchmark, but a new representation layer: learned pattern annotations over simulation traces.

In PHYRE, the agent places a red ball in a cell of an $ 8 \times 8 $ grid with one of three radii, with the goal of causing the green and blue objects to touch. In I-PHYRE, the agent removes objects at different times in the simulation in order to cause the red object to fall out of the scene. In PoolTool, the agent selects shot parameters for striking the cue ball with the goal of potting the 9 ball. In each case, the LLM is given the context about the environment in a prompt (see Appendix E) and prompted to select an action.
We refer readers to their paper for further details. 

DeepPHY uses vision-language models to which videos of simulation rollouts are state inputs. Instead, we provide only the initial image of the simulation along with annotations obtained via our learned pattern library. That is, the state information for simulation roll-out is encoded by our ASTs. We evaluate the effectiveness of the models at selecting actions to solve tasks in the benchmarks. Because preliminary PoolTool performance approached 100\% with the maximum of 15 attempts used in the original paper, we limited all evaluations to 10 attempts.

\begin{table}[t]
    \centering
    \small
    \caption{\F{DeepPHY benchmark success rates (\%) across multiple Qwen model scales. Bold values have one-standard-deviation intervals that overlap with the best result in each environment.}}
    \label{tab:model-scale-results}
    \resizebox{\columnwidth}{!}{%
    \begin{tabular}{llccc}
        \textbf{Model} & \textbf{Feedback} & \textbf{PHYRE} & \textbf{I-PHYRE} & \textbf{PoolTool} \\
        \midrule
        Qwen-3.5  & Image Only        & 8.11 $\pm$ 4.35    & 21.10 $\pm$ 3.29  & 33.30 $\pm$ 5.34 \\
        2B        & Human-Labels      & 9.32 $\pm$ 4.32    & 36.13 $\pm$ 3.13  & 69.96 $\pm$ 7.58 \\
                  & LLM-Labels        & 9.66 $\pm$ 5.13    & 35.89 $\pm$ 3.42  & 68.47 $\pm$ 6.67 \\
        \midrule
        Qwen-3.5  & Image Only        & 8.45 $\pm$ 4.14    & 27.54 $\pm$ 4.13  & 35.66 $\pm$ 3.21 \\
        4B        & Human-Labels      & 14.3 $\pm$ 4.73    & 39.26 $\pm$ 2.04  & 71.47 $\pm$ 5.58 \\
                  & LLM-Labels        & 14.0 $\pm$ 5.21    & 34.22 $\pm$ 3.54  & 74.56 $\pm$ 3.44 \\
        \midrule
        Qwen-3.5  & Image Only        & 9.11  $\pm$ 5.18   & 25.50 $\pm$ 3.91  & 36.23 $\pm$ 4.14 \\
        9B        & Human-Labels      & 14.57 $\pm$ 5.26   & 40.26 $\pm$ 5.26  & 74.45 $\pm$ 6.29 \\
                  & LLM-Labels        & 14.73 $\pm$ 4.89   & 37.55 $\pm$ 4.25  & 75.80 $\pm$ 4.19 \\
        \midrule
        Qwen-3.6  & Image Only        & 13.42 $\pm$ 2.34   & 40.03 $\pm$ 5.23  & 45.81 $\pm$ 5.25 \\
        35B-A3B   & Human-Labels      & \textbf{21.94 $\pm$ 2.13}   & \textbf{54.53 $\pm$ 2.48}  & 80.67 $\pm$ 3.12 \\
                  & LLM-Labels        & \textbf{22.42 $\pm$ 1.16}   & 45.29 $\pm$ 3.16  & 80.36 $\pm$ 2.49 \\
        \midrule
        Qwen-3.6 & Image Only         & 13.23 $\pm$ 3.13   & 45.44 $\pm$ 2.46  & 49.37 $\pm$ 7.36 \\
        27B      & Human-Labels       & \textbf{22.20 $\pm$ 3.11}   & \textbf{55.50 $\pm$ 1.52}  & \textbf{86.47 $\pm$ 3.95} \\
                 & LLM-Labels         & \textbf{22.72 $\pm$ 1.35}   & \textbf{51.58 $\pm$ 4.33}  & \textbf{87.36 $\pm$ 2.49} \\
    \end{tabular}%
    }
\end{table}

\subsection{Human and LLM Labels}\label{sec:human_vs_llm}
For each environment we compare results from using user-supplied labels with LLM-suggested labels. The user labels for patterns are chosen based on relevance for physical reasoning: \texttt{bounce} in PHYRE, \texttt{spring pull towards} in I-PHYRE, \texttt{cushion rebound} in PoolTool, etc. The LLM-suggested labels were obtained by providing the LLM example simulation traces and prompting it to identify 12 relevant pattern labels. Table~\ref{tab:model-scale-results} reports performance across the evaluated models and three environments when using learned pattern annotations as feedback, compared with image-based feedback, i.e., the DeepPHY baseline method. Using the pattern library always results in improved performance relative to the baseline. However, the preference between human and LLM labels depended on the environment and specific labels provided. Our conclusion is that domain experts might be able to tune performance, but LLMs can also be effective at suggesting useful labels.

\begin{figure}[t]
    \centering
    \includegraphics[width=.8\linewidth]{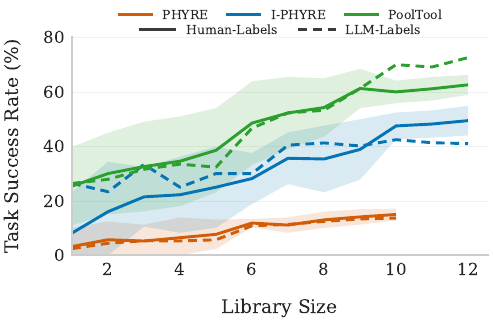}
    \caption{Increasing library size improves performance on DeepPHY benchmarks. The average success rates are plotted against the number of patterns included in the library for human-label and LLM-label libraries. Std. dev. is shaded for the former and omitted (similar magnitudes) for the latter for clarity.}
    
    \label{fig:library_size_results}
\end{figure}

\subsection{Effect of library size}\label{sec:library_size}

We measured the performance in each environment for different sizes of the pattern library. We obtained this by evaluating held-out versions of the ensemble libraries on a reduced benchmark setting, using half of the tasks and half of the maximum number of attempts per task. We repeated the evaluation with progressively larger groups of patterns removed from the library. In each instance, we performed multiple trials. Figure \ref{fig:library_size_results} shows the results of these evaluations for each environment and library variant. As expected, the success rates of tasks in all environments generally improve with library size. We chose to spend our computational budget evaluating all three environments with many trials rather than pushing one environment to a large pattern library. In I-PHYRE and PoolTool environments, we found that a modest size of only 12 patterns was sufficient to achieve 50\%-75\% success rates, while PHYRE struggled to reach 20\% success. This suggests that the optimal library size may depend on the environment and task complexity. We reached diminishing returns with the PHYRE library at around 10 patterns, suggesting that for more difficult environments, pattern choice may be more important than library size.

\begin{figure}[t]
    \centering
    \includegraphics[width=.8\linewidth]{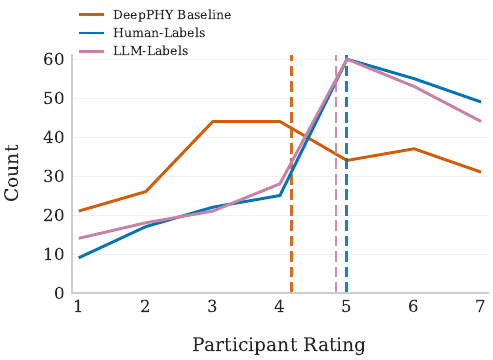}
    \caption{Human evaluators rated summaries generated using learned patterns higher than the DeepPHY baseline.}
    
    \label{fig:survey}
\end{figure}

\subsection{Human evaluation of summaries}\label{sec:user_study}

We conducted a human study on the Prolific \cite{prolific} platform. Participants evaluated three different summaries generated by LMs. They were obtained by using images only, initial image plus annotated patterns (human-label) or the initial image plus annotated patterns (LLM-label). All 100 participants viewed rollout videos from each environment paired with a single summary and rated the summary on a 1--7 Likert scale according to how accurately it describes the video. Summaries were sampled uniformly from the three settings above. Full details and example summaries are provided in Appendix D. Figure \ref{fig:survey} shows that summaries generated using either human-label patterns or LLM-learned patterns are rated more accurate than summaries generated from image frames alone. Thus pattern annotations provide useful high-level information that helps language models produce better descriptions of simulation behaviour. As observed in the benchmark analysis, the gap between human- and LLM-specified labels was insignificant. Another advantage of our approach is the reduced time and number of tokens required to generate summaries (see Table~\ref{tab:summary_stats}). LLM inference was performed on a single NVIDIA A100 40GB GPU and images are at a resolution of $512\times512$ pixels.

\subsection{Evolution of detector programs through learning}

\begin{figure}[t]
    \centering
    \includegraphics[width=.8\linewidth]{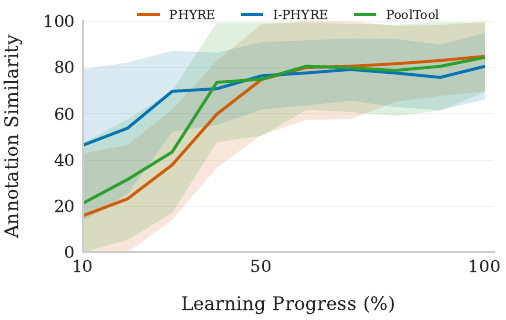}
    \caption{The output from learned code, measured by annotation similarity, smoothly approaches that of the final detectors over training.}
    
    \label{fig:code_convergence}
\end{figure}

To study whether detector synthesis converges toward stable solutions, we measure the similarity of detector behaviour across the course of learning. We use the same annotation-similarity metric employed during evolutionary search, and compare detectors sampled from earlier phases of training to detectors sampled from the final phase. We bin training iterations into 10 bins, each covering $ 10\% $ of the total search budget, and plot the mean and standard deviation of the pattern-similarity between detectors in each bin with those in the final bin. This is measured on the held-out test split of the detector-learning dataset, which contains approximately 100 simulation traces per environment. Figure \ref{fig:code_convergence} shows that, across environments, annotation similarity generally increases over time, indicating that the search process converges toward more stable detector behaviour. The experiment shows that the detector code converges early with diminishing returns after about midway. 

% for appendix:
% - more examples
% - library statistics
% - survey details
% - time + token analysis

 \subsection{Application: Natural language goals via optimisation} \label{sec:applications}
We optimised synthesized reward programs on scenes and measured success rates using hand-coded verification. We used the PoolTool test set of 100 scenes and created 10 natural language goals such as: ``pot the green ball into the top-right pocket'', ``bounce the cue ball off two cushions then pot the orange ball'', or ``knock the green ball into the red ball and pot the red ball''. For each prompt, we synthesized a reward program and optimised the action parameters using simulated annealing. We tested with increasing annealing samples $N_o$ of $ 50 $, $ 100 $, $ 150 $, \dots, $ 1000 $. We ran simulated annealing to select the highest-scoring candidate action under the reward program and executed it in the simulator. To measure success, we wrote verification code for each NL goal using the PoolTool package. We repeated this experiment over multiple random seeds and reported the average success across $N_o$.

\begin{figure}[t]
    \centering
    \includegraphics[width=.8\linewidth]{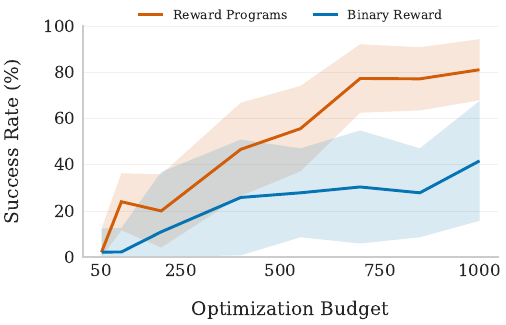}
    \caption{Optimising actions for synthesized reward programs leads to higher success rates on natural language goals compared to optimising sparse binary rewards. The average success rate across 10 natural language goals on 100 held-out scenes is plotted against the optimisation budget. Shaded regions indicate standard deviation across random seeds.}
    
    \label{fig:opt_results}
\end{figure}

We compare our synthesized reward programs, against sparse binary reward functions that return a reward of $ 1 $ if the goal is achieved and $ 0 $ otherwise, and we run the same simulated annealing procedure for both reward types. Figure \ref{fig:opt_results} shows an improvement in optimisation success rates when using synthesized reward programs versus binary rewards, across all optimisation budgets. This trend continues, showing the sample efficiency benefits of using synthesized reward programs that provide dense feedback during optimisation. 

\begin{table}[h]
    \centering
    \footnotesize
    \caption{Our method performs summarization faster and with fewer tokens.}
    \label{tab:summary_stats}
    \resizebox{\columnwidth}{!}{%
    \begin{tabular}{lcccccc}
    \hline
    & \multicolumn{2}{c}{\textbf{PHYRE}} & \multicolumn{2}{c}{\textbf{I-PHYRE}} & \multicolumn{2}{c}{\textbf{PoolTool}} \\
    & \textbf{seconds} & \textbf{tokens} & \textbf{seconds} & \textbf{tokens} & \textbf{seconds} & \textbf{tokens} \\
    \hline
    DeepPHY Baseline     & 85.87 & 5505 & 74.18 & 5505 & 75.72 & 5505 \\
    Human-Labels & 45.84 & 1754 & 38.11 & 984  & 47.97 & 964  \\
    LLM-Labels   & 44.83 & 1720 & 37.71 & 972  & 43.20 & 926  \\
    \hline
    \end{tabular}%
    }
\end{table}

\section{Discussion}

\paragraph{Ensemble library ablation.}\label{sec:ensemble_ablation}
The ensemble library, which groups code with diverse pattern activating behaviour and learns reliability scores, provides a more stable and robust set of annotations for downstream tasks. We perform an ablation of using a single library (without the ensemble optimisation) by evaluating the best performing piece of code for each label from the code pool, across multiple runs. Despite this direct approach not benefiting from noise reduction via ensemble optimisation, the results in Table~\ref{tab:deepphy-results-ablation} show that the simpler approach is still viable. The main difference is in the standard deviation compared to the values in Table~\ref{tab:model-scale-results}, which improves downstream performance. 

\begin{table}[h]
    \centering
    \small
    \caption{Success rates (\%) for single-program (-SP suffix) libraries.}
    \label{tab:deepphy-results-ablation}
    \begin{tabular}{lccc}
        \toprule
         & \textbf{PHYRE} & \textbf{I-PHYRE} & \textbf{PoolTool} \\
        \midrule
        Human-Labels-SP      & 19.61 $\pm$ 2.66 & 51.32 $\pm$ 3.44 & 76.13 $\pm$ 8.36 \\
        LLM-Labels-SP        & 21.52 $\pm$ 1.23 & 39.64 $\pm$ 3.91 & 79.23 $\pm$ 3.3 \\
        \bottomrule
    \end{tabular}
\end{table}

\paragraph{\F{Codex-authored pattern libraries.}}
\F{As an additional robustness check, we evaluated pattern libraries authored by OpenAI GPT-5.6 Sol with the Codex coding agent~\cite{openai_codex}. We measured the performance of these libraries on the PoolTool benchmark using Qwen3.6 35B A3B. Across three independently evaluated Codex-authored libraries, the mean success rate was 73.35\% (72--74\%), as opposed to the 80.67\% mean achieved by Human-Label libraries. These results suggest that a library of patterns can be authored by a coding agent, however the performance is effectively improved by the evolutionary search and ensemble optimisation used in our method, even when using a small local model. The Codex-authored libraries were not optimised for diversity or reliability, which may explain the lower performance.} 

\paragraph{Patterns as helpful abstractions.}
By translating simulation data into annotations of high-level patterns, learned libraries enable LMs to focus on salient physical interactions, improving their ability to select actions in complex physics environments (Figure~\ref{fig:library_size_results}). The pattern library serves as a structured interface that bridges the gap between raw simulation data and LM reasoning capabilities and the resulting annotations complement image data to improve downstream tasks such as summarization and reward synthesis.

\paragraph{Patterns for summarization.}
The results of the human survey (Figure~\ref{fig:survey}) show that the learned pattern library provides annotations that are more helpful for summarization than raw image data alone. Thus, the patterns capture salient physical interactions that are relevant to the dynamics of the environment. The fact that LLM-generated annotations were rated just as accurate as human-generated annotations shows that our method can be effective autonomously, without the need for human labeling. This important result reassures us that learned pattern libraries can provide a scalable way to generate useful annotations for summarization tasks in physics environments.

\paragraph{Annotations enable effective reward synthesis.}
Figure \ref{fig:opt_results} confirms that synthesized reward programs both (i) capture the intended goals specified in natural language, and (ii) provide dense feedback that supports sample-efficient optimisation. These rewards can be easily adapted via natural language through LM refinement, making them a flexible tool for specifying goals. Our results were gathered from 100 random scenes and 10 natural language goals, showing that synthesized rewards are effective across diverse settings.

\paragraph{\F{Common pattern-learning errors.}}
\F{We observed several recurring errors in candidate detector programs. Uncalibrated thresholds, such as requiring cue-ball speed above $0.2$ in PoolTool, produced nearly constant outputs in many candidates. Emitting persistent rolling or spinning states at every timestep produced, in one case, output-count variance above 14,000 and fitness of $-0.956$. High fitness candidates usually remove these errors, but not always: one PoolTool candidate had a $0.358$ correlation score despite retaining fixed thresholds and timestep-level state emission. I-PHYRE frames may be captured at actions, intervals, or termination, so adjacent frames are not uniformly spaced. Some detectors mistook adjacent displacement or reward changes for velocity, energy loss, or reward spikes, whereas better candidates used contact onset, direction reversals, and local extrema. Overall, fitness-based selection favours detectors that distinguish between traces but a high score does not guarantee that detections are physically calibrated or structurally valid. Thus, fitness is useful for ranking candidates, but should not be treated as a complete measure of detector quality.}

\paragraph{Limitations and future work}
Although our pattern discovery method succeeds in forming an abstract, compressed, and interpretable representation for grounded verbal reasoning about physics environments, we acknowledge several limitations. First, learned patterns may trigger noisily in position or timing. Despite this imprecision, our results show improvements over the current state of the art on benchmark tasks, while our ensemble library partly alleviates noisy detections by grouping patterns and learning reliability scores. \F{Second, our trace-distance metric is not invariant to translation or rotation. Consequently, physically equivalent events at different positions or orientations may be treated as distinct, reducing training efficiency and requiring multiple variants of the same semantic pattern. Nevertheless, detectors can use features such as velocity and contact normals to identify patterns across orientations. Future work could incorporate invariant features, canonical coordinate frames, or symmetry-aware distance functions. Third, the learned pattern libraries are environment-specific: our experiments demonstrate the portability of the learning pipeline across environments, rather than direct transfer of learned patterns between them. Future work could investigate shared abstractions and adaptation methods for transferring detectors across related environments. Finally, we allocated our computational budget to robust evaluation, including standard deviations, using small pattern libraries of 12 patterns. Evaluating larger libraries may incur substantial computational and token costs.}

\section{Conclusion}
Our results demonstrate that learning pattern libraries from simulation traces provides a practical interface between physics environments and language models. Across summarization and physics reasoning tasks, these annotations improve LM performance in multiple environments. Finally, by grounding reward program synthesis in the learned pattern library, we enable executable goal specifications from natural language that supports optimisation of complex actions.

\begin{acks}
This work was funded by the UKRI Centre for Doctoral Training in Natural Language Processing (grant EP/S022481/1). Thanks to Kevin Denamganaï and Anna Kapron-King for discussions and feedback in the early stages of this work. We also thank the anonymous reviewers for their insightful comments and suggestions.
\end{acks}

\clearpage
\bibliographystyle{ACM-Reference-Format}
\bibliography{main}

@inproceedings{10.1145/3721238.3730611,
  author = {Chen, Bohong and Li, Yumeng and Zheng, Youyi and Ding, Yao-Xiang and Zhou, Kun},
  title = {Motion-example-controlled Co-speech Gesture Generation Leveraging Large Language Models},
  year = {2025},
  isbn = {9798400715402},
  publisher = {Association for Computing Machinery},
  address = {New York, NY, USA},
  url = {https://doi.org/10.1145/3721238.3730611},
  doi = {10.1145/3721238.3730611},
  booktitle = {Proceedings of the Special Interest Group on Computer Graphics and Interactive Techniques Conference Conference Papers},
  articleno = {55},
  numpages = {12},
  location = {
  },
  series = {SIGGRAPH Conference Papers '25}
}

@inproceedings{10.1145/3641519.3657516,
  author = {Ji, Xuebo and Pan, Zherong and Gao, Xifeng and Pan, Jia},
  title = {Text-Guided Synthesis of Crowd Animation},
  year = {2024},
  isbn = {9798400705250},
  publisher = {Association for Computing Machinery},
  address = {New York, NY, USA},
  url = {https://doi.org/10.1145/3641519.3657516},
  doi = {10.1145/3641519.3657516},
  booktitle = {ACM SIGGRAPH 2024 Conference Papers},
  articleno = {105},
  numpages = {11},
  location = {Denver, CO, USA},
  series = {SIGGRAPH '24}
}

@inproceedings{10.1145/3641519.3657422,
  author = {Sun, Haowen and Zheng, Ruikun and Huang, Haibin and Ma, Chongyang and Huang, Hui and Hu, Ruizhen},
  title = {LGTM: Local-to-Global Text-Driven Human Motion Diffusion Model},
  year = {2024},
  isbn = {9798400705250},
  publisher = {Association for Computing Machinery},
  address = {New York, NY, USA},
  url = {https://doi.org/10.1145/3641519.3657422},
  doi = {10.1145/3641519.3657422},
  booktitle = {ACM SIGGRAPH 2024 Conference Papers},
  articleno = {66},
  numpages = {9},
  location = {Denver, CO, USA},
  series = {SIGGRAPH '24}
}

@inproceedings{10.1145/3641519.3657447,
  author = {Goel, Purvi and Wang, Kuan-Chieh and Liu, C. Karen and Fatahalian, Kayvon},
  title = {Iterative Motion Editing with Natural Language},
  year = {2024},
  isbn = {9798400705250},
  publisher = {Association for Computing Machinery},
  address = {New York, NY, USA},
  url = {https://doi.org/10.1145/3641519.3657447},
  doi = {10.1145/3641519.3657447},
  booktitle = {ACM SIGGRAPH 2024 Conference Papers},
  articleno = {71},
  numpages = {9},
  location = {Denver, CO, USA},
  series = {SIGGRAPH '24}
}

@inproceedings{10.1145/3588432.3591552,
  author = {Gao, William and Aigerman, Noam and Groueix, Thibault and Kim, Vova and Hanocka, Rana},
  title = {TextDeformer: Geometry Manipulation using Text Guidance},
  year = {2023},
  isbn = {9798400701597},
  publisher = {Association for Computing Machinery},
  address = {New York, NY, USA},
  url = {https://doi.org/10.1145/3588432.3591552},
  doi = {10.1145/3588432.3591552},
  booktitle = {ACM SIGGRAPH 2023 Conference Proceedings},
  articleno = {82},
  numpages = {11},
  location = {Los Angeles, CA, USA},
  series = {SIGGRAPH '23}
}

@article{10.1145/3592094,
  author = {Zhang, Longwen and Qiu, Qiwei and Lin, Hongyang and Zhang, Qixuan and Shi, Cheng and Yang, Wei and Shi, Ye and Yang, Sibei and Xu, Lan and Yu, Jingyi},
  title = {DreamFace: Progressive Generation of Animatable 3D Faces under Text Guidance},
  year = {2023},
  issue_date = {August 2023},
  publisher = {Association for Computing Machinery},
  address = {New York, NY, USA},
  volume = {42},
  number = {4},
  issn = {0730-0301},
  url = {https://doi.org/10.1145/3592094},
  doi = {10.1145/3592094},
  journal = {ACM Trans. Graph.},
  month = jul,
  articleno = {138},
  numpages = {16}
}

@article{10.1145/3635705,
  author = {Menapace, Willi and Siarohin, Aliaksandr and Lathuili\`{e}re, St\'{e}phane and Achlioptas, Panos and Golyanik, Vladislav and Tulyakov, Sergey and Ricci, Elisa},
  title = {Promptable Game Models: Text-guided Game Simulation via Masked Diffusion Models},
  year = {2024},
  issue_date = {April 2024},
  publisher = {Association for Computing Machinery},
  address = {New York, NY, USA},
  volume = {43},
  number = {2},
  issn = {0730-0301},
  url = {https://doi.org/10.1145/3635705},
  doi = {10.1145/3635705},
  journal = {ACM Trans. Graph.},
  month = jan,
  articleno = {17},
  numpages = {16}
}

@article{10.1145/3731209,
  author = {Ma, Jiaju and Agrawala, Maneesh},
  title = {MoVer: Motion Verification for Motion Graphics Animations},
  year = {2025},
  issue_date = {August 2025},
  publisher = {Association for Computing Machinery},
  address = {New York, NY, USA},
  volume = {44},
  number = {4},
  issn = {0730-0301},
  url = {https://doi.org/10.1145/3731209},
  doi = {10.1145/3731209},
  journal = {ACM Trans. Graph.},
  month = jul,
  articleno = {33},
  numpages = {17}
}

@inproceedings{ng2000irl,
  title={Algorithms for Inverse Reinforcement Learning},
  author={Ng, Andrew Y. and Russell, Stuart},
  booktitle={Proceedings of the Seventeenth International Conference on Machine Learning},
  pages={663--670},
  publisher={Morgan Kaufmann Publishers Inc.},
  address={San Francisco, CA, USA},
  year={2000}
}

@inproceedings{NEURIPS2018_74934548,
 author = {Vazquez-Chanlatte, Marcell and Jha, Susmit and Tiwari, Ashish and Ho, Mark K and Seshia, Sanjit},
 booktitle = {Advances in Neural Information Processing Systems},
 editor = {S. Bengio and H. Wallach and H. Larochelle and K. Grauman and N. Cesa-Bianchi and R. Garnett},
 pages = {},
 publisher = {Curran Associates, Inc.},
 address = {Red Hook, NY, USA},
 title = {Learning Task Specifications from Demonstrations},
 url = {https://proceedings.neurips.cc/paper_files/paper/2018/file/74934548253bcab8490ebd74afed7031-Paper.pdf},
 volume = {31},
 year = {2018}
}

@misc{prolific,
  author       = {{Prolific}},
  title        = {Prolific},
  howpublished = {\url{https://www.prolific.com/}},
  year         = {2026},
}

@article{toro2018reward,
  title={Reward Machines: Exploiting Reward Function Structure in Reinforcement Learning},
  author={Toro Icarte, Rodrigo and Klassen, Toryn Q. and Valenzano, Richard and McIlraith, Sheila A.},
  journal={Journal of Artificial Intelligence Research},
  volume={73},
  pages={173--208},
  year={2022},
  doi={10.1613/jair.1.12440}
}

@inproceedings{NEURIPS2019_532435c4,
 author = {Toro Icarte, Rodrigo and Waldie, Ethan and Klassen, Toryn and Valenzano, Rick and Castro, Margarita and McIlraith, Sheila},
 booktitle = {Advances in Neural Information Processing Systems},
 editor = {H. Wallach and H. Larochelle and A. Beygelzimer and F. d\textquotesingle Alch\'{e}-Buc and E. Fox and R. Garnett},
 pages = {},
 publisher = {Curran Associates, Inc.},
 address = {Red Hook, NY, USA},
 title = {Learning Reward Machines for Partially Observable Reinforcement Learning},
 url = {https://proceedings.neurips.cc/paper_files/paper/2019/file/532435c44bec236b471a47a88d63513d-Paper.pdf},
 volume = {32},
 year = {2019}
}

@article{Zhou_Li_2022, 
  title={Programmatic Reward Design by Example}, 
  volume={36}, url={https://ojs.aaai.org/index.php/AAAI/article/view/20910}, 
  DOI={10.1609/aaai.v36i8.20910}, 
  number={8}, 
  journal={Proceedings of the AAAI Conference on Artificial Intelligence}, 
  author={Zhou, Weichao and Li, Wenchao}, year={2022}, month={Jun.}, 
  pages={9233-9241} 
}

@misc{liuEvolutionHeuristicsEfficient2024,
  title = {Evolution of {{Heuristics}}: {{Towards Efficient Automatic Algorithm Design Using Large Language Model}}},
  shorttitle = {Evolution of {{Heuristics}}},
  author = {Liu, Fei and Tong, Xialiang and Yuan, Mingxuan and Lin, Xi and Luo, Fu and Wang, Zhenkun and Lu, Zhichao and Zhang, Qingfu},
  year = 2024,
  month = jun,
  number = {arXiv:2401.02051},
  eprint = {2401.02051},
  primaryclass = {cs},
  publisher = {arXiv},
  urldate = {2024-08-27},
  archiveprefix = {arXiv},
  langid = {english}
}

@misc{bai2025qwen3vltechnicalreport,
      title={Qwen3-VL Technical Report}, 
      author={Shuai Bai and Yuxuan Cai and Ruizhe Chen and Keqin Chen and Xionghui Chen and Zesen Cheng and Lianghao Deng and Wei Ding and Chang Gao and Chunjiang Ge and Wenbin Ge and Zhifang Guo and Qidong Huang and Jie Huang and Fei Huang and Binyuan Hui and Shutong Jiang and Zhaohai Li and Mingsheng Li and Mei Li and Kaixin Li and Zicheng Lin and Junyang Lin and Xuejing Liu and Jiawei Liu and Chenglong Liu and Yang Liu and Dayiheng Liu and Shixuan Liu and Dunjie Lu and Ruilin Luo and Chenxu Lv and Rui Men and Lingchen Meng and Xuancheng Ren and Xingzhang Ren and Sibo Song and Yuchong Sun and Jun Tang and Jianhong Tu and Jianqiang Wan and Peng Wang and Pengfei Wang and Qiuyue Wang and Yuxuan Wang and Tianbao Xie and Yiheng Xu and Haiyang Xu and Jin Xu and Zhibo Yang and Mingkun Yang and Jianxin Yang and An Yang and Bowen Yu and Fei Zhang and Hang Zhang and Xi Zhang and Bo Zheng and Humen Zhong and Jingren Zhou and Fan Zhou and Jing Zhou and Yuanzhi Zhu and Ke Zhu},
      year={2025},
      eprint={2511.21631},
      archivePrefix={arXiv},
      primaryClass={cs.CV},
      url={https://arxiv.org/abs/2511.21631}, 
}

@article{romera-paredesMathematicalDiscoveriesProgram2024,
  title = {Mathematical Discoveries from Program Search with Large Language Models},
  author = {{Romera-Paredes}, Bernardino and Barekatain, Mohammadamin and Novikov, Alexander and Balog, Matej and Kumar, M. Pawan and Dupont, Emilien and Ruiz, Francisco J. R. and Ellenberg, Jordan S. and Wang, Pengming and Fawzi, Omar and Kohli, Pushmeet and Fawzi, Alhussein},
  year = 2024,
  month = jan,
  journal = {Nature},
  volume = {625},
  number = {7995},
  pages = {468--475},
  issn = {0028-0836, 1476-4687},
  doi = {10.1038/s41586-023-06924-6},
  urldate = {2024-08-27},
  langid = {english}
}

@misc{yeReEvoLargeLanguage2024a,
  title = {{{ReEvo}}: {{Large Language Models}} as {{Hyper-Heuristics}} with {{Reflective Evolution}}},
  shorttitle = {{{ReEvo}}},
  author = {Ye, Haoran and Wang, Jiarui and Cao, Zhiguang and Berto, Federico and Hua, Chuanbo and Kim, Haeyeon and Park, Jinkyoo and Song, Guojie},
  year = 2024,
  publisher = {arXiv},
  eprint = {2402.01145},
  archivePrefix = {arXiv},
  primaryClass = {cs.NE},
  doi = {10.48550/ARXIV.2402.01145},
  urldate = {2026-01-06},
  copyright = {arXiv.org perpetual, non-exclusive license},
}

@article{battagliaSimulationEnginePhysical2013,
  title = {Simulation as an Engine of Physical Scene Understanding},
  author = {Battaglia, Peter W. and Hamrick, Jessica B. and Tenenbaum, Joshua B.},
  year = 2013,
  month = nov,
  journal = {Proceedings of the National Academy of Sciences},
  volume = {110},
  number = {45},
  pages = {18327--18332},
  issn = {0027-8424, 1091-6490},
  doi = {10.1073/pnas.1306572110},
  urldate = {2026-01-06},
  langid = {english}
}

@misc{bordesIntPhys2Benchmarking2025,
  title = {{{IntPhys}} 2: {{Benchmarking Intuitive Physics Understanding In Complex Synthetic Environments}}},
  shorttitle = {{{IntPhys}} 2},
  author = {Bordes, Florian and Garrido, Quentin and Kao, Justine T and Williams, Adina and Rabbat, Michael and Dupoux, Emmanuel},
  year = 2025,
  publisher = {arXiv},
  eprint = {2506.09849},
  archivePrefix = {arXiv},
  primaryClass = {cs.CV},
  doi = {10.48550/ARXIV.2506.09849},
  urldate = {2026-01-06},
  copyright = {arXiv.org perpetual, non-exclusive license}
}

@misc{bredisEnhancingVisionLanguageModel2025,
  title = {Enhancing {{Vision-Language Model Training}} with {{Reinforcement Learning}} in {{Synthetic Worlds}} for {{Real-World Success}}},
  author = {Bredis, George and Dereka, Stanislav and Sinii, Viacheslav and Rakhimov, Ruslan and Gavrilov, Daniil},
  year = 2025,
  month = aug,
  number = {arXiv:2508.04280},
  eprint = {2508.04280},
  primaryclass = {cs},
  publisher = {arXiv},
  doi = {10.48550/arXiv.2508.04280},
  urldate = {2025-11-07},
  archiveprefix = {arXiv},
  langid = {english}
}

@misc{cherianLLMPhyComplexPhysical2024a,
  title = {{{LLMPhy}}: {{Complex Physical Reasoning Using Large Language Models}} and {{World Models}}},
  shorttitle = {{{LLMPhy}}},
  author = {Cherian, Anoop and Corcodel, Radu and Jain, Siddarth and Romeres, Diego},
  year = 2024,
  publisher = {arXiv},
  eprint = {2411.08027},
  archivePrefix = {arXiv},
  primaryClass = {cs.LG},
  doi = {10.48550/ARXIV.2411.08027},
  urldate = {2026-01-06},
  copyright = {arXiv.org perpetual, non-exclusive license}
}

@misc{assran2025vjepa2selfsupervisedvideo,
      title={V-JEPA 2: Self-Supervised Video Models Enable Understanding, Prediction and Planning}, 
      author={Mido Assran and Adrien Bardes and David Fan and Quentin Garrido and Russell Howes and Mojtaba Komeili and Matthew Muckley and Ammar Rizvi and Claire Roberts and Koustuv Sinha and Artem Zholus and Sergio Arnaud and Abha Gejji and Ada Martin and Francois Robert Hogan and Daniel Dugas and Piotr Bojanowski and Vasil Khalidov and Patrick Labatut and Francisco Massa and Marc Szafraniec and Kapil Krishnakumar and Yong Li and Xiaodong Ma and Sarath Chandar and Franziska Meier and Yann LeCun and Michael Rabbat and Nicolas Ballas},
      year={2025},
      eprint={2506.09985},
      archivePrefix={arXiv},
      primaryClass={cs.AI},
      url={https://arxiv.org/abs/2506.09985}, 
}

@misc{garridoIntuitivePhysicsUnderstanding2025a,
  title = {Intuitive Physics Understanding Emerges from Self-Supervised Pretraining on Natural Videos},
  author = {Garrido, Quentin and Ballas, Nicolas and Assran, Mahmoud and Bardes, Adrien and Najman, Laurent and Rabbat, Michael and Dupoux, Emmanuel and LeCun, Yann},
  year = 2025,
  publisher = {arXiv},
  eprint = {2502.11831},
  archivePrefix = {arXiv},
  primaryClass = {cs.CV},
  doi = {10.48550/ARXIV.2502.11831},
  urldate = {2026-01-06},
  copyright = {Creative Commons Attribution 4.0 International}
}

@misc{kambhampatiLLMsCantPlan2024,
  title = {{{LLMs Can}}'t {{Plan}}, {{But Can Help Planning}} in {{LLM-Modulo Frameworks}}},
  author = {Kambhampati, Subbarao and Valmeekam, Karthik and Guan, Lin and Verma, Mudit and Stechly, Kaya and Bhambri, Siddhant and Saldyt, Lucas and Murthy, Anil},
  year = 2024,
  publisher = {arXiv},
  eprint = {2402.01817},
  archivePrefix = {arXiv},
  primaryClass = {cs.AI},
  doi = {10.48550/ARXIV.2402.01817},
  urldate = {2026-01-06},
  copyright = {Creative Commons Attribution 4.0 International}
}

@misc{liuMindsEyeGrounded2022a,
  title = {Mind's {{Eye}}: {{Grounded Language Model Reasoning}} through {{Simulation}}},
  shorttitle = {Mind's {{Eye}}},
  author = {Liu, Ruibo and Wei, Jason and Gu, Shixiang Shane and Wu, Te-Yen and Vosoughi, Soroush and Cui, Claire and Zhou, Denny and Dai, Andrew M.},
  year = 2022,
  publisher = {arXiv},
  eprint = {2210.05359},
  archivePrefix = {arXiv},
  primaryClass = {cs.CL},
  doi = {10.48550/ARXIV.2210.05359},
  urldate = {2026-01-06},
  copyright = {Creative Commons Attribution Non Commercial Share Alike 4.0 International}
}

@misc{mecattafLittleLessConversation2024,
  title = {A Little Less Conversation, a Little More Action, Please: {{Investigating}} the Physical Common-Sense of {{LLMs}} in a {{3D}} Embodied Environment},
  shorttitle = {A Little Less Conversation, a Little More Action, Please},
  author = {Mecattaf, Matteo G. and Slater, Ben and Te{\v s}i{\'c}, Marko and Prunty, Jonathan and Voudouris, Konstantinos and Cheke, Lucy G.},
  year = 2024,
  publisher = {arXiv},
  eprint = {2410.23242},
  archivePrefix = {arXiv},
  primaryClass = {cs.AI},
  doi = {10.48550/ARXIV.2410.23242},
  urldate = {2026-01-06},
  copyright = {Creative Commons Attribution Share Alike 4.0 International}
}

@misc{paglieriBALROGBenchmarkingAgentic2024,
  title = {{{BALROG}}: {{Benchmarking Agentic LLM}} and {{VLM Reasoning On Games}}},
  shorttitle = {{{BALROG}}},
  author = {Paglieri, Davide and Cupia{\l}, Bart{\l}omiej and Coward, Samuel and Piterbarg, Ulyana and Wolczyk, Maciej and Khan, Akbir and Pignatelli, Eduardo and Kuci{\'n}ski, {\L}ukasz and Pinto, Lerrel and Fergus, Rob and Foerster, Jakob Nicolaus and {Parker-Holder}, Jack and Rockt{\"a}schel, Tim},
  year = 2024,
  month = nov,
  number = {arXiv:2411.13543},
  eprint = {2411.13543},
  publisher = {arXiv},
  urldate = {2024-11-21},
  archiveprefix = {arXiv}
}

@misc{xiang2025aligningperceptionreasoningmodeling,
    title={Aligning Perception, Reasoning, Modeling and Interaction: A Survey on Physical AI}, 
    author={Kun Xiang and Terry Jingchen Zhang and Yinya Huang and Jixi He and Zirong Liu and Yueling Tang and Ruizhe Zhou and Lijing Luo and Youpeng Wen and Xiuwei Chen and Bingqian Lin and Jianhua Han and Hang Xu and Hanhui Li and Bin Dong and Xiaodan Liang},
    year={2025},
    eprint={2510.04978},
    archivePrefix={arXiv},
    primaryClass={cs.AI},
    url={https://arxiv.org/abs/2510.04978}, 
}

@misc{bakhtin2019phyrenewbenchmarkphysical,
    title={PHYRE: A New Benchmark for Physical Reasoning}, 
    author={Anton Bakhtin and Laurens van der Maaten and Justin Johnson and Laura Gustafson and Ross Girshick},
    year={2019},
    eprint={1908.05656},
    archivePrefix={arXiv},
    primaryClass={cs.LG},
    url={https://arxiv.org/abs/1908.05656}, 
}

@misc{qiuPHYBenchHolisticEvaluation2025,
  title = {{{PHYBench}}: {{Holistic Evaluation}} of {{Physical Perception}} and {{Reasoning}} in {{Large Language Models}}},
  shorttitle = {{{PHYBench}}},
  author = {Qiu, Shi and Guo, Shaoyang and Song, Zhuo-Yang and Sun, Yunbo and Cai, Zeyu and Wei, Jiashen and Luo, Tianyu and Yin, Yixuan and Zhang, Haoxu and Hu, Yi and Wang, Chenyang and Tang, Chencheng and Chang, Haoling and Liu, Qi and Zhou, Ziheng and Zhang, Tianyu and Zhang, Jingtian and Liu, Zhangyi and Li, Minghao and Zhang, Yuku and Jing, Boxuan and Yin, Xianqi and Ren, Yutong and Fu, Zizhuo and Wang, Weike and Tian, Xu and Lv, Anqi and Man, Laifu and Li, Jianxiang and Tao, Feiyu and Sun, Qihua and Liang, Zhou and Mu, Yu-Song and Li, Zhong-wei and Zhang, Jing-Jun and Zhang, Shutao and Li, Xiaotian and Xia, Xingqi and Lin, Jiawei and Shen, Zheyu and Chen, Jiahang and Xiong, Qi and Wang, Binran and Wang, Fengyuan and Ni, Ziyang and Zhang, Bohan and Cui, Fan and Shao, Changkun and Cao, Qing-Hong and Luo, Mingjian and Zhang, Muhan and Zhu, Hua Xing},
  year = 2025,
  month = apr,
  eprint = {2504.16074},
  archivePrefix = {arXiv},
  primaryClass = {cs.CL},
  url = {https://arxiv.org/abs/2504.16074},
  urldate = {2025-05-07}
}

@misc{shivanjassimGRASPNovelBenchmark2023,
  title = {{{GRASP}}: {{A}} Novel Benchmark for Evaluating Language {{GRounding And Situated}}   {{Physics}} Understanding in Multimodal Language Models},
  author = {Jassim, Serwan and Holubar, Mario and Richter, Annika and Wolff, Cornelius and Ohmer, Xenia and Bruni, Elia},
  year = 2023,
  month = nov,
  eprint = {2311.09048},
  archivePrefix = {arXiv},
  primaryClass = {cs.CL},
  url = {https://arxiv.org/abs/2311.09048},
  doi = {10.48550/arxiv.2311.09048},
}

@misc{xuDeepPHYBenchmarkingAgentic2025,
  title = {{{DeepPHY}}: {{Benchmarking Agentic VLMs}} on {{Physical Reasoning}}},
  shorttitle = {{{DeepPHY}}},
  author = {Xu, Xinrun and Bu, Pi and Wang, Ye and Karlsson, B{\"o}rje F. and Wang, Ziming and Song, Tengtao and Zhu, Qi and Song, Jun and Ding, Zhiming and Zheng, Bo},
  year = 2025,
  month = aug,
  number = {arXiv:2508.05405},
  eprint = {2508.05405},
  primaryclass = {cs},
  publisher = {arXiv},
  doi = {10.48550/arXiv.2508.05405},
  urldate = {2025-11-07},
  archiveprefix = {arXiv},
  langid = {english}
}

@inproceedings{cuetip_paper2025,
author = {Memery, Sean and Denamgana\"{\i}, Kevin and Zhang, Jiaxin and Tu, Zehai and Guo, Yiwen and Subr, Kartic},
title = {CueTip: An Interactive and Explainable Physics-aware Pool Assistant},
year = {2025},
isbn = {9798400715402},
publisher = {Association for Computing Machinery},
address = {New York, NY, USA},
url = {https://doi.org/10.1145/3721238.3730742},
doi = {10.1145/3721238.3730742},
booktitle = {Proceedings of the Special Interest Group on Computer Graphics and Interactive Techniques Conference Conference Papers},
articleno = {86},
numpages = {11},
location = {
},
series = {SIGGRAPH Conference Papers '25}
}

@misc{memery2024simlmlanguagemodelsinfer,
      title={SimLM: Can Language Models Infer Parameters of Physical Systems?}, 
      author={Sean Memery and Mirella Lapata and Kartic Subr},
      year={2024},
      eprint={2312.14215},
      archivePrefix={arXiv},
      primaryClass={cs.CL},
      url={https://arxiv.org/abs/2312.14215}, 
}

@misc{daineseGeneratingCodeWorld2024,
  title = {Generating {{Code World Models}} with {{Large Language Models Guided}} by {{Monte Carlo Tree Search}}},
  author = {Dainese, Nicola and Merler, Matteo and Alakuijala, Minttu and Marttinen, Pekka},
  year = 2024,
  month = may,
  number = {arXiv:2405.15383},
  eprint = {2405.15383},
  primaryclass = {cs},
  publisher = {arXiv},
  urldate = {2024-08-27},
  archiveprefix = {arXiv},
  langid = {english}
}

@misc{Xu2025PhySensePP,
  title={PhySense: Principle-Based Physics Reasoning Benchmarking for Large Language Models},
  author={Yinggan Xu and Yue Liu and Zhiqiang Gao and Changnan Peng and Di Luo},
  year={2025},
  eprint={2505.24823},
  archivePrefix={arXiv},
  primaryClass={cs.LG},
  url={https://api.semanticscholar.org/CorpusID:279070750}
}

@misc{Chow2025PhysBenchBA,
  title={PhysBench: Benchmarking and Enhancing Vision-Language Models for Physical World Understanding},
  author={Wei Chow and Jiageng Mao and Boyi Li and Daniel Seita and Vitor Campanholo Guizilini and Yue Wang},
  year={2025},
  eprint={2501.16411},
  archivePrefix={arXiv},
  primaryClass={cs.CV},
  url={https://api.semanticscholar.org/CorpusID:275932005}
}

@misc{Xiang2025SeePhysDS,
  title={SeePhys: Does Seeing Help Thinking? - Benchmarking Vision-Based Physics Reasoning},
  author={Kun Xiang and Heng Li and Terry Jingchen Zhang and Yinya Huang and Zirong Liu and Peixin Qu and Jixi He and Jiaqi Chen and Yu-Jie Yuan and Jianhua Han and Hang Xu and Hanhui Li and Mrinmaya Sachan and Xiaodan Liang},
  year={2025},
  eprint={2505.19099},
  archivePrefix={arXiv},
  primaryClass={cs.AI},
  url={https://api.semanticscholar.org/CorpusID:278905129}
}

@inproceedings{Zhang2025PhysReasonAC,
  title={PhysReason: A Comprehensive Benchmark towards Physics-Based Reasoning},
  author={Xinyu Zhang and Yuxuan Dong and Yanrui Wu and Jiaxing Huang and Chengyou Jia and Basura Fernando and Mike Zheng Shou and Lingling Zhang and Jun Liu},
  booktitle={Proceedings of the 63rd Annual Meeting of the Association for Computational Linguistics (Volume 1: Long Papers)},
  pages={16593--16615},
  publisher={Association for Computational Linguistics},
  address={Vienna, Austria},
  year={2025},
  url={https://api.semanticscholar.org/CorpusID:276422440}
}

@article{Davidson_2025,
   title={Goals as reward-producing programs},
   volume={7},
   ISSN={2522-5839},
   url={http://dx.doi.org/10.1038/s42256-025-00981-4},
   DOI={10.1038/s42256-025-00981-4},
   number={2},
   journal={Nature Machine Intelligence},
   publisher={Springer Science and Business Media LLC},
   author={Davidson, Guy and Todd, Graham and Togelius, Julian and Gureckis, Todd M. and Lake, Brenden M.},
   year={2025},
   month=feb, pages={205–220} 
}

@misc{wong2025modelingopenworldcognitionondemand,
      title={Modeling Open-World Cognition as On-Demand Synthesis of Probabilistic Models}, 
      author={Lionel Wong and Katherine M. Collins and Lance Ying and Cedegao E. Zhang and Adrian Weller and Tobias Gerstenberg and Timothy O'Donnell and Alexander K. Lew and Jacob D. Andreas and Joshua B. Tenenbaum and Tyler Brooke-Wilson},
      year={2025},
      eprint={2507.12547},
      archivePrefix={arXiv},
      primaryClass={cs.CL},
      url={https://arxiv.org/abs/2507.12547}, 
}

@misc{ahmed2025synthesizingworldmodelsbilevel,
      title={Synthesizing world models for bilevel planning}, 
      author={Zergham Ahmed and Joshua B. Tenenbaum and Christopher J. Bates and Samuel J. Gershman},
      year={2025},
      eprint={2503.20124},
      archivePrefix={arXiv},
      primaryClass={cs.AI},
      url={https://arxiv.org/abs/2503.20124}, 
}

@misc{curtis2025llmguidedprobabilisticprograminduction,
      title={LLM-Guided Probabilistic Program Induction for POMDP Model Estimation}, 
      author={Aidan Curtis and Hao Tang and Thiago Veloso and Kevin Ellis and Joshua Tenenbaum and Tomás Lozano-Pérez and Leslie Pack Kaelbling},
      year={2025},
      eprint={2505.02216},
      archivePrefix={arXiv},
      primaryClass={cs.AI},
      url={https://arxiv.org/abs/2505.02216}, 
}

@misc{yu2023languagerewardsroboticskill,
      title={Language to Rewards for Robotic Skill Synthesis}, 
      author={Wenhao Yu and Nimrod Gileadi and Chuyuan Fu and Sean Kirmani and Kuang-Huei Lee and Montse Gonzalez Arenas and Hao-Tien Lewis Chiang and Tom Erez and Leonard Hasenclever and Jan Humplik and Brian Ichter and Ted Xiao and Peng Xu and Andy Zeng and Tingnan Zhang and Nicolas Heess and Dorsa Sadigh and Jie Tan and Yuval Tassa and Fei Xia},
      year={2023},
      eprint={2306.08647},
      archivePrefix={arXiv},
      primaryClass={cs.RO},
      url={https://arxiv.org/abs/2306.08647}, 
}

@misc{Dai2025PhysicsArenaTF,
  title={PhysicsArena: The First Multimodal Physics Reasoning Benchmark Exploring Variable, Process, and Solution Dimensions},
  author={Song Dai and Yibo Yan and Jiamin Su and Dongfang Zihao and Yubo Gao and Yonghua Hei and Jungang Li and Junyan Zhang and Sicheng Tao and Zhuoran Gao and Xuming Hu},
  year={2025},
  eprint={2505.15472},
  archivePrefix={arXiv},
  primaryClass={cs.CL},
  url={https://api.semanticscholar.org/CorpusID:278783124}
}

@misc{tang2024worldcodermodelbasedllmagent,
      title={WorldCoder, a Model-Based LLM Agent: Building World Models by Writing Code and Interacting with the Environment}, 
      author={Hao Tang and Darren Key and Kevin Ellis},
      year={2024},
      eprint={2402.12275},
      archivePrefix={arXiv},
      primaryClass={cs.AI},
      url={https://arxiv.org/abs/2402.12275}, 
}

@article{Kiefl2024,
    doi = {10.21105/joss.07301},
    url = {https://doi.org/10.21105/joss.07301},
    year = {2024},
    publisher = {The Open Journal},
    volume = {9},
    number = {101},
    pages = {7301},
    author = {Evan Kiefl},
    title = {Pooltool: A Python package for realistic billiards simulation},
    journal = {Journal of Open Source Software}
}

@misc{li2024iphyreinteractivephysicalreasoning,
      title={I-PHYRE: Interactive Physical Reasoning}, 
      author={Shiqian Li and Kewen Wu and Chi Zhang and Yixin Zhu},
      year={2024},
      eprint={2312.03009},
      archivePrefix={arXiv},
      primaryClass={cs.AI},
      url={https://arxiv.org/abs/2312.03009}, 
}

@misc{qwen36_35b_a3b,
    title = {{Qwen3.6-35B-A3B}: Agentic Coding Power, Now Open to All},
    url = {https://qwen.ai/blog?id=qwen3.6-35b-a3b},
    author = {{Qwen Team}},
    month = {April},
    year = {2026}
}

@misc{openai_codex,
  author = {{OpenAI}},
  title = {Introducing {Codex}},
  year = {2025},
  month = {May},
  howpublished = {\url{https://openai.com/index/introducing-codex/}},
}

@misc{llamacpp,
  author = {{ggml-org}},
  title = {llama.cpp},
  year = {2026},
  publisher = {GitHub},
  journal = {GitHub repository},
  howpublished = {\url{https://github.com/ggml-org/llama.cpp}},
}

@misc{Ma2023EurekaHR,
  title={Eureka: Human-Level Reward Design via Coding Large Language Models},
  author={Yecheng Jason Ma and William Liang and Guanzhi Wang and De-An Huang and Osbert Bastani and Dinesh Jayaraman and Yuke Zhu and Linxi Fan and Anima Anandkumar},
  year={2023},
  eprint={2310.12931},
  archivePrefix={arXiv},
  primaryClass={cs.RO},
  url={https://api.semanticscholar.org/CorpusID:264306288}
}
\clearpage
\appendix
\def\arxivpaper{}

\ifdefined\arxivpaper
\else
\title{Discovering High Level Patterns from Simulation Traces: Supplemental Material}
\author{Sean Memery}
\email{s.memery@ed.ac.uk}
\affiliation{%
  \institution{University of Edinburgh}
  \city{Edinburgh}
  \country{United Kingdom}
}
\author{Kartic Subr}
\email{k.subr@ed.ac.uk}
\orcid{0000-0002-7302-4383}
\affiliation{%
  \institution{University of Edinburgh}
  \city{Edinburgh}
  \country{United Kingdom}
}
\renewcommand\shortauthors{Memery and Subr}
\onecolumn
\begin{center}
  {\LARGE Discovering High Level Patterns from Simulation Traces: Supplemental Material\par}
  \vspace{1em}
  {\large Sean Memery and Kartic Subr\par}
  {\normalsize University of Edinburgh\par}
\end{center}

\appendix
\fi

\section{Pattern Learning Statistics}
\begin{table*}[h]
    \centering
    \begin{tabular}{lccccccc}
    \hline
    & \textbf{\# Candidates} & \textbf{\% Executable} & \textbf{Best fitness} & \textbf{Avg fitness} & \textbf{Fitness std dev} & \textbf{Avg LOC} & \textbf{Activation \%}  \\
    \hline
    PHYRE    & 5,501.0 & 99.9\% & 0.2135  &  0.0184  & 0.1913 & 116.8  & 65.9\%  \\
    IPHYRE   & 5,395.2 & 99.9\%  & 0.2125 & -0.0466  & 0.1385 & 137.3 & 56.0\%  \\
    POOLTOOL & 4,737.0 & 99.1\%  & 0.2946 & -0.0556  & 0.1602 & 81.6  & 65.0\%  \\
    \hline
    \end{tabular}
    \caption{Candidate generation and fitness statistics across environments.}
\end{table*}

\section{Full list of learned patterns.}

\noindent\textbf{PHYRE human-labels.}
\begin{lstlisting}[escapechar=@]
moving object hits stationary object
near collision
support relationship
lose support
falling object
rolling object
sliding contact
airborne motion
bounce
grid cell transition
wedged
tip over
\end{lstlisting}

\noindent\textbf{I-PHYRE human-labels.}
\begin{lstlisting}[escapechar=@]
collision
support contact
falling object
swinging object
object removal
sliding contact
airborne motion
bounce
chain movement
spring pull towards
spring resist movement
lever launch
\end{lstlisting}

\noindent\textbf{PoolTool human-labels.}
\begin{lstlisting}[escapechar=@]
cue strike
ball collision
cushion rebound
ball pocketed
rolling ball
sliding ball
spinning ball
ball slowing
spin shot
ball curves around ball
high angle collision
gentle touch
\end{lstlisting}

\noindent\textbf{PHYRE LLM-labels.}
\begin{lstlisting}[escapechar=@]
rest position
ground contact
static barrier
settling state
gravity stability
rest state
static support
object falling
object balance
collision contact
object placement
object support
\end{lstlisting}

\noindent\textbf{I-PHYRE LLM-labels.}
\begin{lstlisting}[escapechar=@]
spring initial tension
support transition
structural release
platform occupancy
bounce
vertical descent
ball landing
gravity drop
support loss
block removal
spring pull
swinging object
\end{lstlisting}

\noindent\textbf{PoolTool LLM-labels.}
\begin{lstlisting}[escapechar=@]
ball slowing
collision
ball pocket
cluster isolation
multibody collision chain
balls clustered
obstructed path
cue ball slide
straight shot line
wide striking angle
cushion rebound
cue ball spin
\end{lstlisting}

\newpage
\section{Example code for a learned pattern}
\noindent\textbf{Example Code: ``Bounce'' Pattern in the I-PHYRE environment.}

\begin{lstlisting}[style=py, caption={}, label={}]
def find_pattern(trace):
    """
    Detect 'bounce' events in an IPHYRE trace.
    """
    patterns = []
    timesteps = trace.get("timesteps", [])
    if len(timesteps) < 3:
        return patterns

    # Identify dynamic objects that are likely to be bouncing
    # We look for objects that are dynamic and have changed position significantly
    dynamic_objects = []
    for ts in timesteps:
        for obj in ts.get("objects", []):
            if obj.get("dynamic", 0) != 0:
                dynamic_objects.append(obj)

    # Group dynamic objects by their index to track individual objects over time
    obj_history = {}
    for ts in timesteps:
        for obj in ts.get("objects", []):
            idx = obj.get("index")
            if idx is not None:
                if idx not in obj_history:
                    obj_history[idx] = []
                obj_history[idx].append((ts.get("index", 0), ts.get("t", 0), obj))

    # For each dynamic object, look for velocity reversal
    for idx, history in obj_history.items():
        if len(history) < 3:
            continue

        # Calculate vertical velocities between consecutive timesteps
        # We look for a pattern: v1 < 0 (falling), v2 > 0 (rising)
        # This indicates a bounce occurred between the timestep where v1 ends and v2 starts.

        velocities = []
        for i in range(1, len(history)):
            prev_t, prev_time, prev_obj = history[i-1]
            curr_t, curr_time, curr_obj = history[i]

            # Calculate center y
            prev_y = (prev_obj.get("y1", 0) + prev_obj.get("y2", 0)) / 2.0
            curr_y = (curr_obj.get("y1", 0) + curr_obj.get("y2", 0)) / 2.0

            dt = curr_time - prev_time
            if dt == 0:
                continue

            v_y = (curr_y - prev_y) / dt
            velocities.append((prev_t, curr_t, v_y, prev_obj, curr_obj))

        # Look for sign change in velocity from negative to positive
        for i in range(1, len(velocities)):
            prev_v = velocities[i-1][2]
            curr_v = velocities[i][2]

            # Check for bounce: previous velocity was negative (falling), current is positive (rising)
            if prev_v < -0.01 and curr_v > 0.01:
                # We found a bounce!
                # The bounce occurred between velocities[i-1] and velocities[i]
                # Start timestep is velocities[i-1][0] (start of the falling phase)
                # End timestep is velocities[i][1] (end of the rising phase)

                start_timestep = velocities[i-1][0]
                end_timestep = velocities[i][1]

                # Get object details
                prev_obj = velocities[i-1][3]
                curr_obj = velocities[i][3]

                # Determine geometry
                shape = "line"
                if prev_obj.get("x1") == prev_obj.get("x2") and prev_obj.get("y1") == prev_obj.get("y2"):
                    shape = "circle"

                # Create object details
                object_details = {
                    f"obj_{idx}": {
                        "role": "bouncing_object",
                        "geometry": shape,
                        "start_position": [prev_obj.get("x1", 0), prev_obj.get("y1", 0)],
                        "end_position": [curr_obj.get("x2", 0), curr_obj.get("y2", 0)]
                    }
                }

                pattern = {
                    "label": "bounce",
                    "start_timestep": start_timestep,
                    "end_timestep": end_timestep,
                    "parameters": {
                        "object_details": object_details
                    }
                }

                patterns.append(pattern)

    return patterns
\end{lstlisting}

\section{Human Survey Details}

\begin{figure*}[t]
    \centering
    \includegraphics[width=\textwidth]{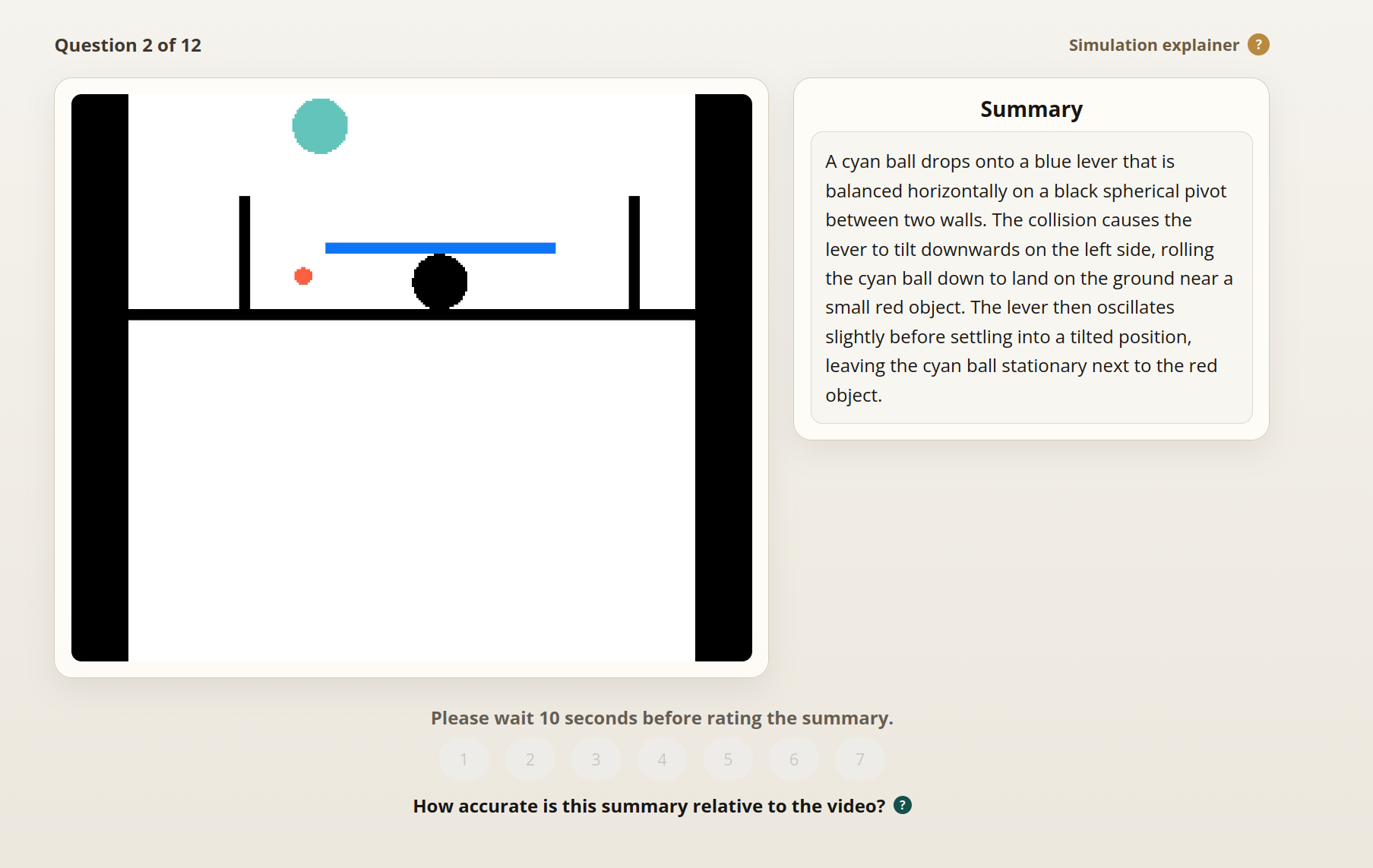}
    \caption{Example of the survey interface shown to human evaluators.}
    \Description{}
    \label{fig:survey_example}
\end{figure*}

\noindent\textbf{Summary example A (Human-Labels).}
\begin{lstlisting}[escapechar=@]
The red ball begins moving south-east after a velocity change, rolling across the green bar and briefly tapping it. This rolling contact propels the green bar toward the blue bar. As the green bar moves, it eventually makes stable contact with the blue bar, forming a persistent stack. The red ball continues its path, interacting with the red jar and black bars through opposing motion and brief taps, but these interactions do not disrupt the green-blue contact.
\end{lstlisting}

\medskip
\noindent\textbf{Summary example B (Image Only).}
\begin{lstlisting}[escapechar=@]
A red ball is launched toward the scene, colliding with the red jar and displacing it. The moving jar then strikes the green bar, propelling it laterally. The green bar slides across the surface, approaching the blue bar. The blue bar adjusts its position slightly as the green bar nears, maintaining a path toward contact. The green bar's continued motion brings its end into direct contact with the blue bar's end. This sustained interaction ensures stable contact at the final frame. The sequence demonstrates a clear causal chain: the red ball's impact initiated the jar's movement, which transferred energy to the green bar, guiding it to meet the blue bar. No obstacles interrupted the motion, allowing precise alignment. The green and blue bars achieved full contact, satisfying the goal with no gap remaining. The outcome confirms the simulation succeeded due to sequential, energy-transfer interactions.
\end{lstlisting}

\medskip
\noindent\textbf{Summary example C (LLM-Labels).}
\begin{lstlisting}[escapechar=@]
The red ball begins moving south-east after a velocity change. The ball's movement triggers a rolling contact with the green bar, briefly supporting it before the green bar settles onto the blue bar. Subsequent interactions stabilize the green bar atop the blue bar, forming a persistent stack.
\end{lstlisting}

\medskip
\noindent\textbf{Survey evaluation help guide.}
\begin{lstlisting}[escapechar=@]
An accurate summary should describe the main events in the video without fabricating details, match the sequence and outcomes shown, and avoid omitting key events. In particular, commonly incorrect details are: colours, object shapes, sequence of events, and positions. 
\end{lstlisting}

\medskip
\noindent\textbf{Survey environment help guide.}
\begin{lstlisting}[escapechar=@]
PHYRE is a 2D physics puzzle environment. The goal is to place a red ball so that the green and blue objects touch. You can pause the video and drag the progress bar to move backward or forward through the simulation before rating the summary.
IPHYRE is a 2D physics puzzle environment. The goal is to get the red ball to fall into the abyss. You can pause the video and drag the progress bar to move backward or forward through the simulation before rating the summary.
PoolTool is a billiards environment. The goal is to pocket coloured balls. You can pause the video and drag the progress bar to move backward or forward through the simulation before rating the summary.
\end{lstlisting}

\section{Language Model Prompts}

Below are all of the prompts used throughout our system (besides some minor prompts for error handling and refining outputs).

\begin{lstlisting}[escapechar=@, caption={Domain Specific Language Specification for LM}, label={lst:dsl_syntax}]
Reward DSL reference (parameter-aware):

- PATTERN("uid", {params?}): true if an event with the given UID/label occurs. Optional params let you require matching event parameters (see library parameter schemas below). Example: PATTERN("abstraction_503681", {"red_ball_id": 8, "green_object_id": 7})
- AND(expr1, expr2, ...): all child expressions must evaluate to true.
- OR(expr1, expr2, ...): returns true when at least one child expression is true.
- NOT(expr): logical negation.
- AFTER("uid_a", "uid_b", min_delta=None, max_delta=None, first_params=None, second_params=None): true if uid_a occurs after uid_b and within optional time bounds; params filter each event.
- WITHIN("uid_a", "uid_b", window, event_params=None, reference_params=None): shorthand for AFTER where uid_a occurs no more than `window` seconds after uid_b.
- COUNT("uid", count, params=None) / GT / LT: count occurrences of an event (optionally filtered by parameters).
- NEARBY_AT(obj_id, x, y, t, threshold_strength=0.1): true if object `obj_id` is within `threshold_strength * 256` units of point (x, y) at simulation time `t` in [0,1]. Use this for spatial proximity checks.
- OBJECT_ID(color, shape): returns the object ID for the object with the given color and shape in the scene (e.g., OBJECT_ID("red", "circle")). Use this to get IDs for NEARBY_AT or PATTERN parameters.

Examples:
    AND(
        PATTERN("abstraction_217640"),
        AFTER("abstraction_217640", "abstraction_307800", first_params={"red_ball_id": 8}, second_params={"green_object_id": 7}),
        NOT(PATTERN("abstraction_612355", {"frame_index": 44}))
    )
    OR(
        COUNT("abstraction_612355", 2, {"red_ball_id": 3}),
        GT("abstraction_661256", 3)
    )
    NEARBY_AT(OBJECT_ID("red", "circle"), 100.0, 150.0, 0.5, threshold_strength=0.2)

IMPORTANT:
- Identifiers are case-insensitive when matching event labels; UIDs are matched exactly.
- Parameter matching requires ALL provided keys to match the emitted event parameters (strings are case-insensitive; numbers match by value). Leave params empty to match any.
- Identifiers, such as NEARBY_AT, should not have their arguments included in the expression, i.e. correct use is NEARBY_AT(5, 100.0, 150.0, 0.5), not NEARBY_AT(obj_id=5, x=100.0, y=150.0, t=0.5).
- Identifiers MUST exist in the library; DO NOT invent new ones.
- See the event library documentation for parameter schemas for built-in and abstraction events.
\end{lstlisting}

\begin{lstlisting}[escapechar=@, caption={LM Reward Program Synthesis Prompt}, label={lst:lm_reward_prompt}]
Given the following goal and the DSL reference, propose one DSL expression. This expression defines a reward function. This reward function will be maximised by an optimisation process. Do not include any comments in the DSL code.
Events can expose parameters; use them to precisely target entities (see library parameter schemas below).

TIPS:
- It is easier for the optimisation process to maximise the reward function when it is expressed as a sum of positive terms i.e. a step by step addition of clauses with AND operators.
- Think about how you want to achieve the goal, and express that in the reward function. Don't just describe a single end condition, but build up the reward function step by step.
- Try to reason about the image that you see, this is the state that the optimisation process will be acting in.
- DO NOT include comments in the DSL code.

Goal:
{{ goal }}

DSL reference:
{{ dsl_guide }}

Library summary:
{{ library_summary }}

Scene summary:
{{ scene_summary }}

Respond in think/answer blocks, with ONLY the DSL inside ```dsl fences.
<think>reason carefully referencing evidence</think>
<answer>```dsl
... your DSL ...
```</answer>

\end{lstlisting}

\begin{lstlisting}[escapechar=@, caption={PHYRE Code Evolution Prompt}, label={lst:phyre_code_evolution_prompt}]
You are writing a Python detector for the {{ environment_id|upper }} environment.
Target primitive label: {{ label }}
Search step: {{ step }}

Write Python code only. Define exactly one function:

def find_pattern(trace):
    ...

Rules:
- no imports
- inspect only the provided `trace` dict
- return a list of pattern dicts
- each pattern dict must contain `label`, `start_timestep`, `end_timestep`, and `parameters`
- use inclusive timestep spans; if the pattern is effectively instantaneous, set `start_timestep == end_timestep`
- choose the narrowest specific timestep span supported by the evidence; do not default to the full trace unless the pattern truly persists for the entire trace
- `parameters` must include an `object_details` dict keyed by object id or role, with each entry describing the object's role in the pattern, shape, color, and where the pattern starts/ends for that object
- include object-location details directly in `parameters` when you can infer them from the trace; do not leave the pattern undescribed if object participation is visible
- optional fields are `provenance` and `metadata`
- use the exact target label `{{ label }}`
- if the pattern is absent, return []

PHYRE trace structure:
- `trace["timesteps"]` is the main signal; it is a dense sequence of simulation frames
- each timestep has `index`, `t`, `state`, `objects`, `actions`, `attachments`
- each object typically has `object_index`, `shape`, `color`, `diameter`, `x`, `y`, `angle`, `vx`, `vy`, `angular_velocity`, `is_user_input`
- the user-controlled placed ball is typically the object with `is_user_input == True`
- scene geometry and target objects are distinguished mainly by `shape` and `color`
- `trace["events"]` is sparse and usually only contains coarse status entries like `simulation_status` or `motion_sequence`
- `trace["outcome"]` holds final task result information such as `success`, `terminal_status`, and metrics like `is_solved`

How to write good PHYRE detectors:
- derive patterns from motion and object relations over time, not just final outcome
- use color/shape semantics when useful, especially green, blue, red, black, purple objects
- detect things like contact, support, containment, escape, or rolling by comparing object positions across timesteps
- prefer simple geometry heuristics over brittle exact thresholds
- do not assume object indices are consistent across different tasks
- when emitting a pattern, mark the inclusive timestep range where it is visibly active
- pick concrete start/end timesteps where the pattern is actually visible, not just the earliest and latest timestep in the trace

Compact trace example:
```python
{
  "environment": "phyre",
  "task_id": "str",
  "timesteps": [
    {
      "index": int,
      "t": float,
      "objects": [
        {
          "object_index": int,
          "shape": "str",
          "color": "str",
          "diameter": float,
          "x": float,
          "y": float,
          "angle": float,
          "vx": float,
          "vy": float,
          "angular_velocity": float,
          "is_user_input": bool
        }
      ]
    }
  ],
  "events": [{"kind": "str", "payload": {...}}],
  "outcome": {"success": bool, "terminal_status": "str", "metrics": {...}}
}
```

Your objective is to maximize the training reward.
Learn from this example what works, keep the strong ideas, remove weak or brittle logic, and produce a better detector.

Improve this parent candidate:
```python
{{ parent_code }}
```

Parent candidate training reward to maximize: {{ parent_metadata.average_score }}
\end{lstlisting}

\begin{lstlisting}[escapechar=@, caption={I-PHYRE Code Evolution Prompt}, label={lst:iphyre_code_evolution_prompt}]
You are writing a Python detector for the {{ environment_id|upper }} environment.
Target primitive label: {{ label }}
Search step: {{ step }}

Write Python code only. Define exactly one function:

def find_pattern(trace):
    ...

Rules:
- no imports
- inspect only the provided `trace` dict
- return a list of pattern dicts
- each pattern dict must contain `label`, `start_timestep`, `end_timestep`, and `parameters`
- use inclusive timestep spans; if the pattern is effectively instantaneous, set `start_timestep == end_timestep`
- choose the narrowest specific timestep span supported by the evidence; do not default to the full trace unless the pattern truly persists for the entire trace
- `parameters` must include an `object_details` dict keyed by object id or role, with each entry describing the object's role in the pattern, geometry/shape, and where the pattern starts/ends for that object
- include object-location details directly in `parameters` when you can infer them from the trace; do not leave the pattern undescribed if object participation is visible
- optional fields are `provenance` and `metadata`
- use the exact target label `{{ label }}`
- if the pattern is absent, return []

I-PHYRE trace structure:
- `trace["timesteps"]` is the main signal
- each timestep has `objects`; each object usually has `index`, `x1`, `y1`, `x2`, `y2`, `radius`, `eliminable`, `dynamic`, `joint`, `spring`
- objects are better interpreted as line segments / endpoints / circular extents than as simple points
- the object center can be approximated from `((x1+x2)/2, (y1+y2)/2)`
- for circular objects, `x1 == x2` and `y1 == y2`
- `joint` and `spring` are numeric identifiers; `0` means none, nonzero values identify a joint/spring group
- `trace["events"]` often contains sparse `scheduled_action` entries that mark external interventions
- `trace["outcome"]` stores task reward, solved threshold, translated actions, and terminal status

How to write good I-PHYRE detectors:
- use geometry over time: support, contact, containment, or escape often appear in relative segment positions
- use `dynamic`, `eliminable`, `joint`, and `spring` values to distinguish object roles
- scheduled actions are informative but usually not sufficient on their own; combine them with state changes when possible
- prefer robust relative comparisons instead of exact coordinates because tasks differ
- emit the inclusive timestep range where the pattern is clearly present
- pick concrete start/end timesteps where the pattern is actually visible, not just the earliest and latest timestep in the trace

Compact trace example:
```python
{
  "environment": "iphyre",
  "task_id": "str",
  "timesteps": [
    {
      "index": int,
      "t": float,
      "objects": [
        {
          "index": int,
          "x1": float,
          "y1": float,
          "x2": float,
          "y2": float,
          "radius": float,
          "eliminable": float,
          "dynamic": float,
          "joint": int,
          "spring": int
        }
      ]
    }
  ],
  "events": [{"kind": "scheduled_action", "t": float, "payload": {...}}],
  "outcome": {"success": bool, "terminal_status": "str", "metrics": {...}}
}
```

Your objective is to maximize the training reward.
Learn from this example what works, keep the strong ideas, remove weak or brittle logic, and produce a better detector.

Improve this parent candidate:
```python
{{ parent_code }}
```

Parent candidate training reward to maximize: {{ parent_metadata.average_score }}
\end{lstlisting}

\begin{lstlisting}[escapechar=@, caption={PoolTool Code Evolution Prompt}, label={lst:pooltool_code_evolution_prompt}]
You are writing a Python detector for the {{ environment_id|upper }} environment.
Target primitive label: {{ label }}
Search step: {{ step }}

Write Python code only. Define exactly one function:

def find_pattern(trace):
    ...

Rules:
- no imports
- inspect only the provided `trace` dict
- return a list of pattern dicts
- each pattern dict must contain `label`, `start_timestep`, `end_timestep`, and `parameters`
- use inclusive timestep spans; if the pattern is effectively instantaneous, set `start_timestep == end_timestep`
- choose the narrowest specific timestep span supported by the evidence; do not default to the full trace unless the pattern truly persists for the entire trace
- `parameters` must include an `object_details` dict keyed by object id or role, with each entry describing the object's role in the pattern, ball identity/kind, and where the pattern starts/ends for that object
- include object-location details directly in `parameters` when you can infer them from the trace; do not leave the pattern undescribed if object participation is visible
- optional fields are `provenance` and `metadata`
- use the exact target label `{{ label }}`
- if the pattern is absent, return []

Pooltool trace structure:
- `trace["timesteps"]` is the main kinematic signal
- timestep objects are usually billiard balls with `kind`, `ball_id`, `position`, `velocity`, `angular_velocity`, `motion_state_code`, `motion_state`, `t`, `radius`
- `position`, `velocity`, and `angular_velocity` are 3D vectors, but table motion is mostly in the x-y plane
- `trace["events"]` is especially informative here; it often contains event kinds such as `physics_ball_ball`, `physics_ball_linear_cushion`, `physics_stick_ball`, `physics_sliding_rolling`, `physics_rolling_stationary`, `shot_prepared`, and `shot_outcome`
- `trace["outcome"]` contains shot-level success and terminal status like potted/no-ball outcomes

How to write good Pooltool detectors:
- use `events` for discrete contact/transition patterns such as collision, rolling transfer, or cue strike
- use `timesteps` when you need sustained motion, support, containment-like geometry, or ball trajectories
- identify balls by `ball_id`, not by list position
- for geometric checks, usually ignore the z coordinate and reason in x-y
- motion-state transitions are often easier to detect than raw acceleration thresholds
- emit inclusive timestep ranges, not a single representative time
- pick concrete start/end timesteps where the pattern is actually visible, not just the earliest and latest timestep in the trace

Compact trace example:
```python
{
  "environment": "pooltool",
  "task_id": "str",
  "timesteps": [
    {
      "index": int,
      "t": float,
      "objects": [
        {
          "kind": "ball",
          "ball_id": "str",
          "position": [float, float, float],
          "velocity": [float, float, float],
          "angular_velocity": [float, float, float],
          "motion_state_code": int,
          "motion_state": "str",
          "radius": float
        }
      ]
    }
  ],
  "events": [{"kind": "physics_ball_ball", "t": float, "payload": {...}}],
  "outcome": {"success": bool, "terminal_status": "str", "metrics": {...}}
}
```

Your objective is to maximize the training reward.
Learn from this example what works, keep the strong ideas, remove weak or brittle logic, and produce a better detector.

Improve this parent candidate:
```python
{{ parent_code }}
```

Parent candidate training reward to maximize: {{ parent_metadata.average_score }}
\end{lstlisting}

\begin{lstlisting}[escapechar=@, caption={LM Label Suggestion Prompt}, label={lst:lm_label_suggestion_prompt}]
You are generating reusable event pattern labels and descriptions for a 2D physics simulation. These patterns will help analysts understand key moments in simulation traces.

Simulation domain (information given to analysts):
{{ trace_spec }}

Objective:
- The ONLY goal is for the green object to touch the blue object.

What patterns are:
- Named, reusable events that help a human visualize key moments in a trace.
- Code will be written later to detect these patterns; your output seeds those detectors.

Current library snapshot (UID, label if any, description):
{{ library_table }}

RL reasoning (<think> snippets that reveal what analysts care about):
{{ rl_thinks }}

Your task:
{{ abstract_guidance }}
- Propose {{ K }} NEW patterns (not duplicates or near-duplicates of existing ones).
- Make sure each pattern is distinct and captures a unique aspect of the simulation traces.
- It is very important that suggested patterns are not similar to existing ones in the library.
- Each pattern:
  - reason: why this pattern is useful given the objective and the reasoning
  - description: one sentence describing the pattern in detail
  - label: a short, descriptive phrase (3-7 words), importantly it should be scene-agnostic

Output:
Make sure to think about what patterns should be suggested first, then output a JSON array like this:
```json
[
  {"reason": "one or two sentences", "description": "one sentence", "label": "short sentence"},
    ...
]
```
\end{lstlisting}

\begin{lstlisting}[escapechar=@, caption={LM Code Repair Prompt}, label={lst:lm_code_repair_prompt}]
Repair the {{ environment_id|upper }} detector candidate for the target label `{{ label }}`.

Return Python code only with exactly one function:

def find_pattern(trace):
    ...

Constraints:
- no imports
- inspect only the provided `trace` dict
- return a list of pattern dicts
- every emitted pattern must include `label`, `start_timestep`, `end_timestep`, and `parameters`
- use inclusive timestep spans; if the pattern is point-like, set `start_timestep == end_timestep`
- choose the narrowest specific timestep span supported by the evidence; do not default to the full trace unless the pattern truly persists for the entire trace
- `parameters` must include an `object_details` dict keyed by object id or role, with each entry describing the object's role in the pattern, ball identity/kind, and where the pattern starts/ends for that object
- use the exact target label `{{ label }}`
- preserve the detector intent unless the validation error requires structural change

{{ environment_id|upper }} reminder:
{{ environment_rules }}

Validation error:
- category={{ error_category }}
- message={{ error_message }}
- line={{ error_line }}
- column={{ error_column }}

Broken candidate:
```python
{{ broken_code }}
```
\end{lstlisting}

\newpage
\section{Code Evolution Details}

\noindent\textbf{FunSearch Algorithm}

Algorithm \ref{alg:funsearch} outlines the FunSearch procedure used for program synthesis via LLMs. The algorithm maintains multiple islands of program candidates, periodically resetting lower-performing islands to promote diversity and exploration. We adapted this method by providing our own Evaluation function tailored to our code-learning task.

\begin{algorithm}[t]
\SetAlgoLined
\KwIn{evaluation function $\mathrm{Evaluate}(\cdot)$, \\
\hspace{2.75em} initial (or skeleton) program $g_0(\cdot)$,  \\
\hspace{2.75em} $\mathrm{LLM}(\cdot)$, \\
\hspace{2.75em} number of islands $I$, prompt size $s$, reset period $T_r$}
\KwOut{Best program found $g^\star(\cdot)$}

\For{$i \leftarrow 1$ \KwTo $I$}{
    $\mathcal{D}_i \leftarrow \{g_0\}$
}
Initialize program database as islands $\mathcal{D} \leftarrow \{\mathcal{D}_i\}_{i=1}^I$\;

\For{iteration $t \leftarrow 1$ \KwTo budget}{
    Sample an island $\mathcal{D}_i$ (favor islands with higher best score)\;
    Sample $s$ best programs $g_1,\dots,g_s$ from $\mathcal{D}_i$ \;
    $g_{new} \gets \mathrm{LLM}(\mathrm{BuildPrompt}(g_1,\dots,g_s))$\;
    \If{$g_{new}$ is valid}{
        $\nu \gets \mathrm{Evaluate}(g_{new})$\;
        Add $(g_{new}, \nu)$ to island $\mathcal{D}_i$\;
    }
    \If{$t \bmod T_r = 0$}{
        Identify worst half of islands\;
        Reinitialize lower half of islands by cloning a top program \; 
    }
}
Return $(g^\star, \nu^\star) \gets \text{argmax}_{(g,\nu)\in \mathcal{D}} \nu$\; % \KS{argmax?}
\caption{FunSearch}
\label{alg:funsearch}
\end{algorithm}

\noindent\textbf{ComputeLengthPenalty}

We apply a logarithmic penalty based on code length. We count the total number of lines and compute the penalty as, $$ \lambda = \log(\texttt{num\_lines})/5 .$$ To avoid unbounded growth, we cap $\ \texttt{num\_lines} \ $ at 1000, which yields a maximum penalty of approximately 1.38.

\noindent\textbf{ComputeTimePenalty}

We use a time-based penalty derived from the average annotation time of existing patterns in the library. Let $\ t \ $ be the average annotation time for the new pattern and let $\ \mu \ $ be the mean annotation time across existing patterns. If $\ t \le \mu \ $, the penalty is 0. If $\ t > \mu \ $, we apply a linear penalty that increases from 0 to 1 as $\ t \ $ rises from $\ \mu \ $ to $\ 2\mu \ $; specifically, the penalty reaches 1 when the new pattern takes twice the mean time. This value is the maximum penalty—any slower pattern (i.e., $\ t \ge 2\mu \ $) receives a penalty of 1.

\newpage
\section{Reward program optimisation}

\noindent\textbf{Natural language goals and DSL reward programs}
\lstset{
  caption={Goals used in the reward program optimisation experiment (natural language paired with DSL)},
  basicstyle=\ttfamily\footnotesize,
  columns=fullflexible
}
\begin{lstlisting}[escapechar=@]
[1] NL: Make a ball curve around another ball, then pot the eight ball.
    DSL: AND(PATTERN("ball curves around ball", ball="cue"), CUE_HIT("8"), AFTER(PATTERN("ball curves around ball", ball="cue"), CUE_HIT("8")), AFTER(CUE_HIT("8"), PATTERN("ball pocket", ball="8")))

[2] NL: Make a rolling ball spin after contact, then pot the black ball.
    DSL: AND(PATTERN("rolling ball", ball="cue"), CUE_HIT("8"), AFTER(PATTERN("rolling ball", ball="cue"), CUE_HIT("8")), AFTER(CUE_HIT("8"), PATTERN("spinning ball", ball="cue")), AFTER(CUE_HIT("8"), PATTERN("ball pocket", ball="8")))

[3] NL: Make a spinning ball curve around another ball and pot the red ball.
    DSL: AND(PATTERN("spinning ball", ball="cue"), PATTERN("ball curves around ball", ball="cue"), CUE_HIT("3"), AFTER(PATTERN("ball curves around ball", ball="cue"), CUE_HIT("3")), AFTER(CUE_HIT("3"), PATTERN("ball pocket", ball="3")))

[4] NL: Make a sliding ball hit the blue ball at a high angle, then pot the blue ball.
    DSL: AND(PATTERN("sliding ball", ball="cue"), CUE_HIT("2"), PATTERN("high angle collision", contains=["cue", "2"]), AFTER(PATTERN("sliding ball", ball="cue"), CUE_HIT("2")), AFTER(CUE_HIT("2"), PATTERN("ball pocket", ball="2")))

[5] NL: Make a ball curve around a blocker, glance off a cushion, and pot the green ball.
    DSL: AND(PATTERN("ball curves around ball", ball="cue"), CUE_HIT("6"), PATTERN("cushion rebound", ball="6"), AFTER(CUE_HIT("6"), PATTERN("cushion rebound", ball="6")), AFTER(PATTERN("cushion rebound", ball="6"), PATTERN("ball pocket", ball="6")))

[6] NL: Make a ball hit the brown ball at a high angle, then have the brown ball pot without touching a cushion.
    DSL: AND(CUE_HIT("7"), PATTERN("high angle collision", contains=["cue", "7"]), AFTER(CUE_HIT("7"), PATTERN("ball pocket", ball="7")), NOT(PATTERN("cushion rebound", ball="7")))

[7] NL: Make a sliding ball bounce off a cushion then hit the red ball at a high angle.
    DSL: AND(PATTERN("sliding ball", ball="cue"), PATTERN("cushion rebound", ball="cue"), AFTER(PATTERN("cushion rebound", ball="cue"), CUE_HIT("3")), AFTER(PATTERN("cushion rebound", ball="cue"), PATTERN("high angle collision", contains=["cue", "3"])))

[8] NL: Make a ball softly touch the blue ball, have the blue ball rebound off a cushion.
    DSL: AND(PATTERN("gentle touch", contains=["cue", "2"]), CUE_HIT("2"), PATTERN("cushion rebound", ball="2"), AFTER(CUE_HIT("2"), PATTERN("cushion rebound", ball="2")))

[9] NL: Make a spinning ball hit the green ball at a high angle, then have the green ball curve around another ball.
    DSL: AND(PATTERN("spinning ball", ball="cue"), CUE_HIT("6"), PATTERN("high angle collision", contains=["cue", "6"]), AFTER(PATTERN("high angle collision", contains=["cue", "6"]), PATTERN("ball curves around ball", ball="6")))

[10] NL: Make a ball rebound from two cushions, then pot the eight ball.
     DSL: AND(PATTERN_COUNT("cushion rebound", ball="cue", min_count=2), AFTER(PATTERN_COUNT("cushion rebound", ball="cue", min_count=2), CUE_HIT("8")), AFTER(CUE_HIT("8"), PATTERN("ball pocket", ball="8")))
\end{lstlisting}

\newpage
\section{Summarising examples}

\begin{figure*}[t]
    \centering
    \includegraphics[width=\textwidth]{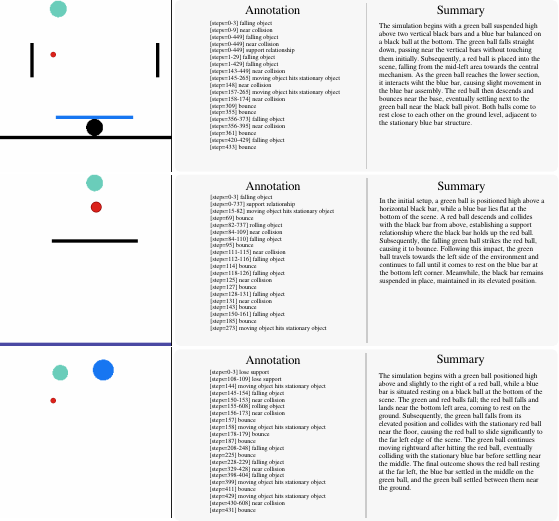}
    \caption{Example summaries of tasks in the PHYRE environment, generated by the human-label ensemble library.}
    \Description{}
    \label{fig:phyre_summaries}
\end{figure*}
\begin{figure*}[t]
    \centering
    \includegraphics[width=\textwidth]{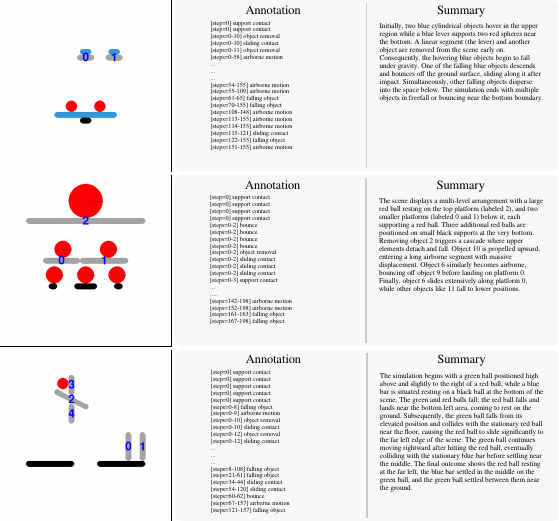}
    \caption{Example summaries of tasks in the I-PHYRE environment, generated by the human-label ensemble library.}
    \Description{}
    \label{fig:iphire_summaries}
\end{figure*}
\begin{figure*}[t]
    \centering
    \includegraphics[width=\textwidth]{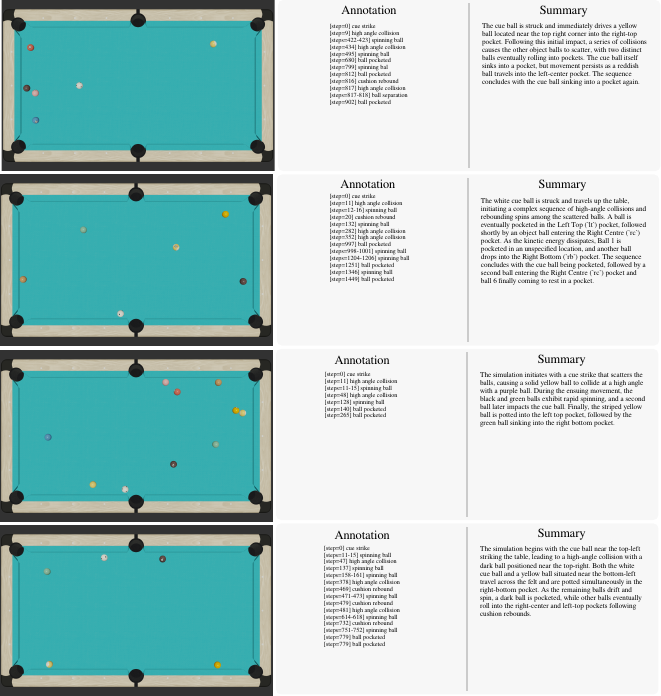}
    \caption{Example summaries of tasks in the PoolTool environment, generated by the human-label ensemble library.}
    \Description{}
    \label{fig:pooltool_summaries}
\end{figure*}

\end{document}